\documentclass[preprint,12pt]{elsarticle}
\usepackage{epsfig}
\usepackage{float,lscape}
\usepackage{rotating}
\usepackage{epstopdf}
\usepackage[ruled,vlined]{algorithm2e}
\usepackage[utf8]{inputenc}
\usepackage{amsmath}
\usepackage{amsfonts}
\usepackage{amssymb}
\usepackage{graphicx}
\usepackage{geometry}
\usepackage{setspace}
\usepackage{lmodern}
\usepackage{longtable}
\usepackage{relsize}
\usepackage{color,soul}
\usepackage{multirow}
\usepackage{subfigure}

\usepackage{enumerate}
\usepackage{lineno,hyperref}
\usepackage{natbib}
\usepackage{pifont}

\journal{Journal}

\begin{document}

\begin{frontmatter}

\title{Direct Speech to Speech Translation: A Review}


\author{$^{1}$Mohammad Sarim}
\author{$^{1}$Saim Shakeel}
\author{$^{1}$Laeeba Javed}
\author{$^{1}$Jamaluddin}
\author{$^{1}$Mohammad Nadeem}

\address{$^{1}$Department of Computer Science, Aligarh Muslim University, Aligarh, Uttar Pradesh, India.}

\begin{abstract}
Speech-to-speech translation (S2ST) is a transformative technology that bridges global communication gaps, enabling real-time multilingual interactions in diplomacy, tourism, and international trade. This review examines the evolution of S2ST, comparing traditional cascade models—which rely on automatic speech recognition (ASR), machine translation (MT), and text-to-speech (TTS) components—with newer end-to-end and direct speech translation (DST) models that bypass intermediate text representations. While cascade models offer modularity and optimized components, they suffer from error propagation, increased latency, and loss of prosody. In contrast, direct S2ST models, such as Translatotron 2 and UnitY, retain speaker identity, reduce latency, and improve translation  naturalness by preserving vocal characteristics and prosody. However, they remain limited by data sparsity, high computational costs, and generalization challenges for low-resource languages. This paper critically evaluates these approaches, their trade-offs, and future directions for improving real-time multilingual communication.
\end{abstract}

\end{frontmatter}
\section{Introduction}

Speech translation is the interpretation of spoken language from one language to another without the loss of meaning and intent. It is more complex than text translation because it processes nuances, intonation, and the rhythm of speech in real time \cite{communication2023seamlessmultilingualexpressivestreaming}. With globalization, there is an increased need for speech translation with a view to making seamless communication across language barriers (Seamless: Multilingual Expressive and Streaming Speech Translation). This is particularly important in the context of international business, diplomacy, and tourism, through which ideas can be exchanged easily and mutual understanding encouraged \cite{jia2022translatotron2highqualitydirect}.

Speech-to-speech translation refers to the direct translation of speech in one language into another language, including voice characteristics of the speaker. More advanced than text translation because it is processed in real-time, with accents, intonations, and even the emotional tone of the original speech \cite{le2024transvipspeechspeechtranslation}. This technology is significant because communication and interaction with people from different linguistic backgrounds have become the norm. In international business, it ensures clear communication, which prevents misunderstandings that could potentially lead to costly mistakes. In diplomatic contexts, it will help convey a nuanced message with clarity to create better relations between nations. It also enhances travel experiences because tourists can interact more freely with locals, enriching cultural exchanges and making the whole travel experience smoother and enjoyable \cite{ijcat}.

Cascade models are somewhat more multi-step speech-to-speech translation processes where first, ASR transforms the input speech into text and then, MT does a follow-up and converts into the target language; then finally, TTS takes this output and transforms into speech. This helps in optimizing each one of the components but yet exploits some techniques for those special areas that ASR, MT, and TTS belong to. Combining such specialized modules, cascade models can make full use of state-of-the-art algorithms for each stage, making speech translation systems with more powerful capabilities \cite{wu2016googlesneuralmachinetranslation} \cite{li-etal-2017-translating}.

Cascade models have another advantage-modularity. That is to say, enhancement in one module will feed forward to enhance the overall system automatically. For example, because of improvements in ASR, there are fewer errors in transcription during the first go-round. That means that it improves the accuracy of translation. As MT can be improved upon further to produce even better translation, TTS can also be improved to make the speech sound better. Hence, a cascade model is quite pliable and absorbs advancements as they come along \cite{app11104380} \cite{dasgupta2020improvinglocalidentifiabilityprobabilistic}.

Cascade models have limitations, though. The primary limitation is error propagation-the inaccuracies introduced in the Automatic Speech Recognition (ASR) stage are forwarded to the Machine translation (MT) stage, and accumulate in the output of the cascade. Additionally, since each stage takes time to process, and must finish before the next can be initiated, there is added latency to the process as well. It requires a lot of effort to include and calibrate isolated modules, and constructing and even maintaining cascade models is very hard and resource intensive. Second, the intermediate text form is often less convenient and leads to some loss in prosodic and contextual information often available for the spoken equivalent. Yet research continues making these models more accurate day by day \cite{jia2019directspeechtospeechtranslationsequencetosequence} \cite{pratapa-etal-2018-language}.

Direct models for speech-to-speech translation (S2ST) work directly to translate spoken language in one language into another, bypassing the intermediate text representation \cite{jia2022translatotron2highqualitydirect}. One of the most prominent representatives of this kind is Translatotron 2. That makes use of an end-to-end neural network, consisting of three components: a speech encoder and a linguistic decoder with an acoustic synthesizer, which has demonstrated remarkable improvements in translation quality and voice preservation in comparison with its predecessor2. Another very important model in the list is SeamlessM4T, a highly multilingual and multimodal machine translation system that was developed to better facilitate unbroken communication between any two languages \cite{communication2023seamlessm4tmassivelymultilingual}.

Direct S2ST models have a few drawbacks. In fact, one major weakness of this technology is that it has insufficient data compared to others; the approach still remains highly dependent on large amounts of parallel speech and text data compared to cascaded systems. The dependency can negatively affect the performance as well as the generalization capability of the direct models \cite{jia2022translatotron2highqualitydirect}, \cite{10.1007/978-3-031-48312-7_21}. This includes linguistic and acoustic diversity, as the speech to be translated must display characteristic features of a range of languages and accents. Furthermore, since the building of the project itself calls for advanced models and heavy computational power, high-quality translations cannot be attained at high decoding speeds \cite{han2024speechqeestimatingqualitydirect}, \cite{xu2023recentadvancesdirectspeechtotext}, \cite{key}.

Important changes have recently been realized in direct S2ST models. Of all models, Translatotron 2 is most outstanding in using a speech encoder, a linguistic decoder, an acoustic synthesizer, and a singular attention module to successfully translate while keeping the speaker's voice \cite{jia2022translatotron2highqualitydirect}. Another novel model that has been recently presented is UnitY based on discrete units for two-pass direct S2ST; it could also enhance both translation quality and robustness \cite{xu2023recentadvancesdirectspeechtotext}. There were more models presented lately. More examples include direct speech-to-speech translation with discrete units that employ self-supervised discrete speech encoders and sequence-to-sequence speech to unit translation models for performance gains \cite{lee2022directspeechtospeechtranslationdiscrete}.

India is the land of over 1,600 languages and dialects. Therefore, some unique special challenges and opportunities India poses in linguistic diversity hold for S2ST. Tremendous variations of modeling exist in the language and acoustics. New promises include the Direct S2ST models such as Translatotron 2 that aimed to exploit the full scope of end-to-end neural networks for converting speech into speech, without mid-term text representation \cite{jia2022translatotron2highqualitydirect}. However, several Indian languages lack parallel speech data \cite{10.1007/978-3-031-48312-7_21}. Researchers of the Indian Institute of Technology Dharwad have thoroughly documented sequence-to-sequence models that use transformers. This tremendous achievement was also fairly evident in Prabhu-pada-vani, which had parallel Hindi/English speech datasets. This work also depicts that direct S2ST models will fill the lacuna of linguistic diversity in India, but it strongly represents that it requires stronger data collection and model training such that it will be effective in meeting the linguistic diversities of the country \cite{han2024speechqeestimatingqualitydirect}, \cite{xu2023recentadvancesdirectspeechtotext}, \cite{key}.

Only very few extensive reviews have been carried out on direct S2ST models. Among the prominent reviews, the existing literature is categorized into three major domains of concern: modeling burden, data scarcity, and application-related challenges. Results of the review show the relevance of end-to-end models in minimizing latency and also that an error does not accumulate in a manner similar to that in cascaded systems. Approaches concerning data scarcity are various, including data augmentation, pre-training, and knowledge distillation \cite{jia2022translatotron2highqualitydirect}, \cite{10.1007/978-3-031-48312-7_21}. \cite{han2024speechqeestimatingqualitydirect}.

\begin{figure*}
    \centering
    \includegraphics[width=\textwidth, height=\textheight, keepaspectratio]{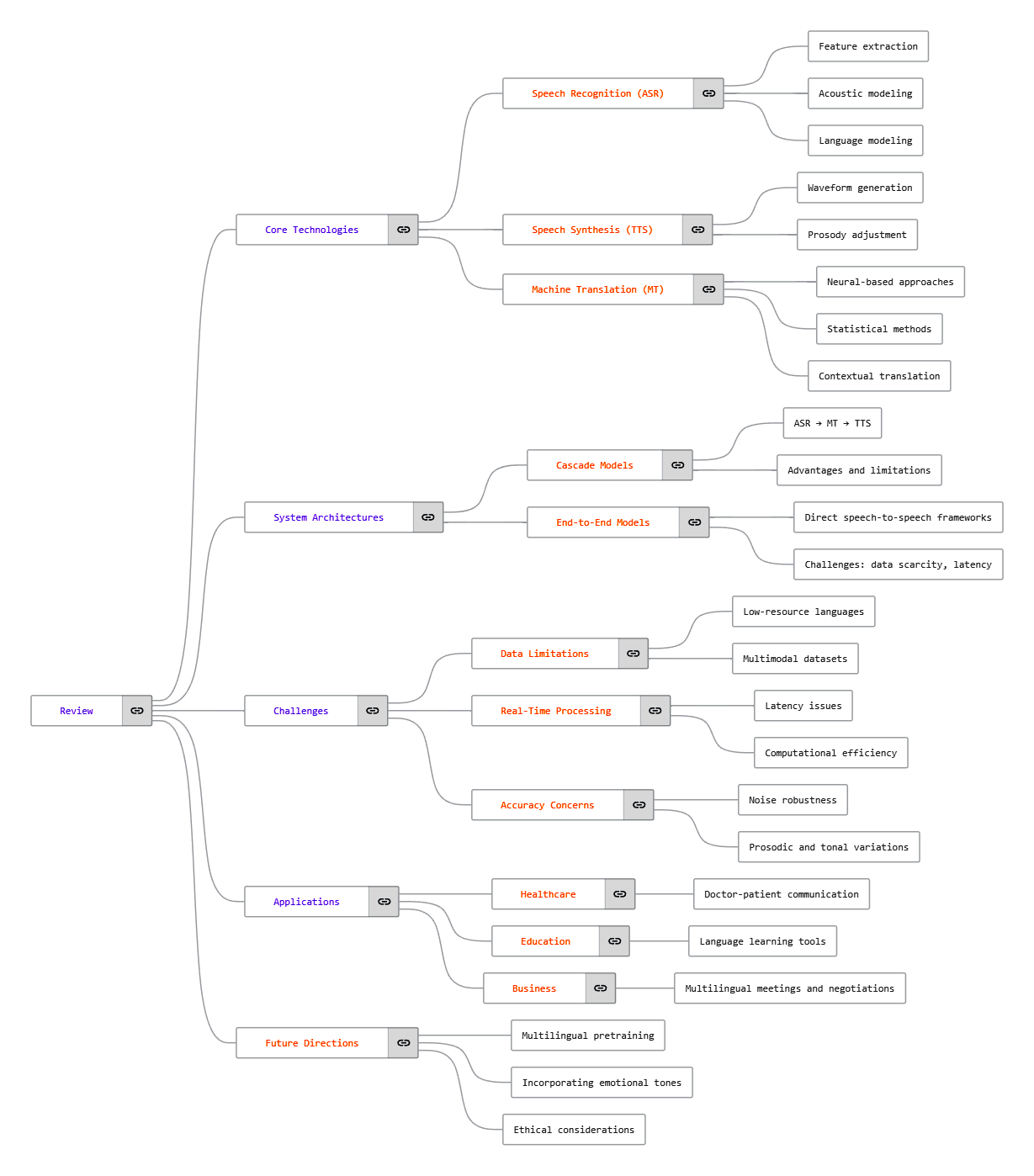}
    \caption{Classifications used in the current study}
    \label{fig:tree}
\end{figure*}

The proposed reviews aim to extend the current direct S2ST model by developing and furthering the understanding of this class of models. One particular proposed review involves a comprehensive review capturing contemporary state-of-the-art methodologies, which split efforts to classify them along the prevalent challenges. Therefore, the proposed review hereby puts emphasis on high-quality data acquisition and model training, especially with linguistic diversity, so that the task can be properly managed. Apart from this, some other possible areas for further research are real-time translation, segmentation, named entity recognition, gender bias, and code-switching \cite{inaguma2023unitytwopassdirectspeechtospeech} \cite{lee2022directspeechtospeechtranslationdiscrete}. On a broader scale, Figure~\ref{fig:tree} provides an overview of the key discussions presented in the proposed paper.

\section{Background}

\subsection{Direct or end-to-end speech translation}
End-to-end speech translation is one of the state-of-the-art approaches, directly bridging spoken language into another spoken language without any intermediate text representation \cite{jia2022translatotron2highqualitydirect}. To be specific, instead of implementing separate stages of speech recognition, text translation, and finally, text-to-speech synthesis as in cascaded systems, end-to-end models encompass the entire process. Error propagation risks resulting from different stages are thereby reduced, and this ensures faster and better-quality translation \cite{GoogleAIBlog}. The conventional architecture of end-to-end speech-to-text translation mainly consists of a speech encoder, a linguistic decoder, an acoustic synthesizer, and an attention mechanism that aligns input and output sequences accurately \cite{wei2022jointpretrainingspeechbilingual}.

The major benefit of this approach is that it can achieve faster inference by bypassing intermediate steps, cutting down on compounded errors, and preserving the speaker's vocal characteristics in the translated speech \cite{Al-Tarawneh2024}. This streamlined process yields faster and more precise translations and is therefore very precious when used in real-time communication. Reduced latency and propagation of errors make end-to-end models more reliable and applicable in practice for tasks like multilingual meetings, live broadcasts, and international customer service interactions \cite{PullumKornai2003}.

The mathematical premise of all of these involves sequence-to-sequence models, usually realized as neural networks \cite{Rahmawati2020}. The attention mechanism becomes a key enabler for lining up input and output sequences to establish a direct mapping from the source to the target speech \cite{GoogleAIBlog}. In model training, the loss function
\textit{L} that encodes the gap between the estimated and real target sequences and can be formulated as:

\begin{align}
\mathcal{L} = -\sum_{t=1}^{T} \log P(y_{t} | y_{<t}, \mathbf{x}) 
\end{align}

where \textit{x} denotes the input sequence and $y_{t}$ denotes the target sequence at time step $t$ \cite{jia2022translatotron2highqualitydirect}. Evaluation is done using metrics such as BLEU scores \cite{10.3115/1073083.1073135}\cite{DBLP:journals/corr/abs-1804-08771}. This new concept of speech translation has proven great potential in improving the accuracy and speed of multilingual communication systems \cite{Al-Tarawneh2024} \cite{Kornai2007}.

\subsection{Related Works}
Speech-to-speech translation (S2ST) systems have undergone tremendous change from one based on pipeline models to direct translation systems. Pipeline models that are based on the ASR-MT-TTS sequence can be seen in \cite{ijcaonline}. Though each system is modular and thus allows for improving individual stages, problems such as latency and error propagation across stages remain. These systems also risk losing prosodic and contextual nuances due to their intermediate processing via text. Noteworthy examples of existing pipelines dedicated to predominantly multilingual communications such as IBM MASTOR and Verbmobil show the limits of a cover with regard to spontaneous speech and low-resource languages.

The direct S2ST systems address a lot of the challenges mentioned previously by eliminating any intermediary text representation. \cite{arxiv1} discursively classified direct S2ST models like Translatotron 2, UnitY, etc., that are based on end-to-end architectures combining speech encoding, linguistic decoding, and acoustic synthesis. This situation favors unwritten or low-resource languages as it allows translation between speech units their own, without textual annotations. Nevertheless, such direct systems face issues related to scarcity of data, computational requirements, and keeping speaker voice characteristics in real time.

\cite{ kaur2024directpunjabienglishspeech} introduces a transformer-based S2ST model that converts a sequence of discrete units from the source language into a sequence of discrete units from the destination language using discrete acoustic units. \cite{fang2024achievehighqualitydirectspeechtospeech} Introduces a dual architecture for translating between French and Fulfulde.\cite{carolefrench}  presents a brand-new S2ST model architecture that combines an arbitrary S2TT and TTS model connector with a CTC-based vocabulary adapter to create an S2ST model. \cite{liu2023multi} Enhances Tibetan-Chinese speech translation using multi-task learning. \cite{nachmani2024translatotron} Builds on previous Translatotron versions, enabling high-quality translation using only monolingual data. \cite{lee2022direct} Utilizes discrete speech units to enhance the direct speech-to-speech translation pipeline. This approach leverages self-supervised learning to encode target speech into discrete units, which are then used for direct translation, bypassing the need for intermediate text generation. \cite{ jia2022translatotron2highqualitydirect} improves upon the original Translatotron by preserving the speaker's voice during translation, addressing performance bottlenecks such as suboptimal utilization of auxiliary textual supervision, challenges in modeling translation alignment, and robustness issues in speech generation. \cite{ zhang2021uwspeech} The paper demonstrates that UWSpeech significantly outperforms direct translation and VQ-VAE baselines, highlighting its potential in effectively translating unwritten languages by leveraging cross-lingual speech recognition and synthesis techniques. \cite{  jia2019direct} introduces an attention-based sequence-to-sequence neural network that directly translates speech from one language to another without intermediate text. Trained end-to-end, it maps source speech spectrograms to target language spectrograms. Comparative analysis of recently developed direct speech translation models are shown in Table~\ref{table:summary}.

\begin{table*}
\centering
\caption{Comparative Study of Recent End-to-End Speech Translation Models}
\resizebox{\textwidth}{!}{%
\scriptsize 
\begin{tabular}{|p{1cm}|p{1cm}|p{2cm}|p{2.5cm}|p{2cm}|p{2.5cm}|}
\hline
\textbf{Year} & \textbf{Paper} & \textbf{Dataset} & \textbf{Model Used} & \textbf{Language Worked Upon} & \textbf{Limitations} \\ \hline

2024 & \cite{kaur2024directpunjabienglishspeech} & FLEURS dataset \& Kathbath & Transformer-based U2UT Model & Punjabi - English & Dataset size and available computing \\ \hline
2024 & \cite{fang2024achievehighqualitydirectspeechtospeech} & Agricultural words recorded by natives & Dual architecture combining direct and cascading methods & French - Fulfulde & Dataset is undersupplied, as the number of people who made word recordings is very insignificant \\ \hline
2024 & \cite{carolefrench} & CVSS & ComSpeech and ComSpeech-ZS & French - English, German - English, Spanish - English & ComSpeech-ZS performance still lags behind the cascaded system where TTS data is scarce \\ \hline
2023 & \cite{liu2023multi} & Tibetan-Chinese parallel speech data & Multi-task self-supervised learning & Chinese - English & Scarcity of parallel corpus data \\ \hline
2023 & \cite{nachmani2024translatotron} & LibriSpeech, Fisher Spanish-English & Translatotron 3 & Spanish - English & Reliance on monolingual data for training may affect translation diversity \\ \hline
2022 & \cite{lee2022direct} & LibriSpeech, Fisher Spanish-English, CVSS-C, multi-domain EnEs corpora including Europarl-ST, MuST-C, TEDLIUM3, Common Voice, CoVoST2, and mTEDx & Discrete unit-based translation model & Spanish - English & Challenges in unit extraction and synthesis, as well as the complexity involved in aligning and generating discrete units accurately \\ \hline
2022 & \cite{jia2022translatotron2highqualitydirect} & LibriSpeech, Fisher Spanish-English & Translatotron 2 & Spanish - English & Complexity in voice preservation and translation accuracy \\ \hline
2021 & \cite{zhang2021uwspeech} & Custom dataset for unwritten languages, including Fisher Spanish-English conversation dataset & Model for unwritten languages using XL-VAE (cross-lingual vector quantized variational autoencoder) & Spanish - English & Limited availability of high-quality data for unwritten languages \\ \hline
2019 & \cite{jia2019direct} & Conversational Spanish-to-English and Fisher Spanish-to-English & Translatotron & Spanish - English & Performance not as good as a baseline cascade of ST and TTS models \\ \hline
\label{table:summary}
\end{tabular}%
}
\end{table*}
\cite{10229412} continues to expand this debate with new approaches in speech-to-unit translation besides self-supervised models, such as wav2vec 2.0 and mHuBERT, that are much less dependent on text data and therefore much better adapted to noisy and accentally diverse settings. It argues for the significance of discrete unit representations, which could allow linguistic variation to be handled in a more robust way without sacrificing prosodic information. \cite{10229412} offers the organized classification of low-resource speech-to-speech translation research trends and implements the task toward semi-supervised learning and data augmentation to enhance linguistic diversity.

Earlier works, well summarized by \cite{igntu}, have largely dealt with STT and TTS in multilingual environments. The relevance of such work to S2ST is mainly through central techniques including feature extraction methods like MFCC and LPC and hybrid models that merge rule-based and statistical methods. Though the strategies mentioned above provided a basis for all later neural architectures, direct translation and real-time capabilities were not underlined specifically.

Our review advances these insights by providing a comprehensive taxonomy of direct S2ST models with special emphasis on multilingual and low-resource applications, especially in linguistically diverse contexts like India. Table~\ref{tab:survey} compares the recent surveys with this study,Unlike other experiments, our study is directed toward a holistic dataset-based approach, unique machine learning models and methods, along with several challenges that one faces in multiple industries due to data availability, how to deal with syntactic differences, and designing optimum model architectures. Future work directions in our work also include increasing language support and improving model performance, reducing the noise robustness, along with introducing both paralinguistic and non-linguistic factors towards improving resilience in various application scenarios.

\begin{table*}
\centering
\caption{Survey of Direct Speech-to-Speech Neural Machine Translation and Related Research}
\resizebox{\textwidth}{!}{%
\scriptsize
\begin{tabular}{|p{3cm}|p{2cm}|p{1cm}|p{1cm}|p{1cm}|p{1cm}|p{1cm}|p{1cm}|p{1cm}|p{1cm}|}
\hline
\textbf{Paper Title} & 
\multicolumn{4}{|c|}{\textbf{Scrutiny of Dataset}} & 
\textbf{Feature Extraction} & 
\textbf{ML Models} & 
\textbf{ML Techniques} & 
\textbf{Future Direction} & 
\textbf{Bibliography Analysis} \\ \hline

& \textbf{Characteristics} & 
  \textbf{Annotations and Preprocessing} & 
  \textbf{Data Sources and Modalities} & 
  \textbf{Limitations and Benchmarks} & 
  & & & & \\ \hline

Direct Speech-to-Speech Neural Machine Translation: A Survey & 
  \centering \ding{51} & 
  \ding{51} & 
  \ding{51} & 
  \ding{51} & 
  \ding{51} & 
  \ding{51} & 
  \ding{51} & 
  \ding{51} & 
  \ding{55} \\ \hline

Research Opportunities in Automatic Speech-to-Speech Translation & 
  \centering \ding{55} & 
  \ding{55} & 
  \ding{55} & 
  \ding{55} & 
  \ding{55} & 
  \ding{51} & 
  \ding{55} & 
  \ding{55} & 
  \ding{51} \\ \hline

A Comprehensive Survey on Automatic Speech Recognition Using Neural Networks & 
  \centering \ding{55} & 
  \ding{55} & 
  \ding{51} & 
  \ding{55} & 
  \ding{51} & 
  \ding{55} & 
  \ding{55} & 
  \ding{51} & 
  \ding{51} \\ \hline

Proposed study & 
 \centering \ding{51} & 
  \ding{51} & 
  \ding{51} & 
  \ding{51} & 
  \ding{51} & 
  \ding{51} & 
  \ding{51} & 
  \ding{51} & 
  \ding{51} \\ \hline
\end{tabular}%
}
\label{tab:survey}
\end{table*}

\section{Methodology}
\begin{figure*}
    \centering
    \includegraphics[width=\textwidth]{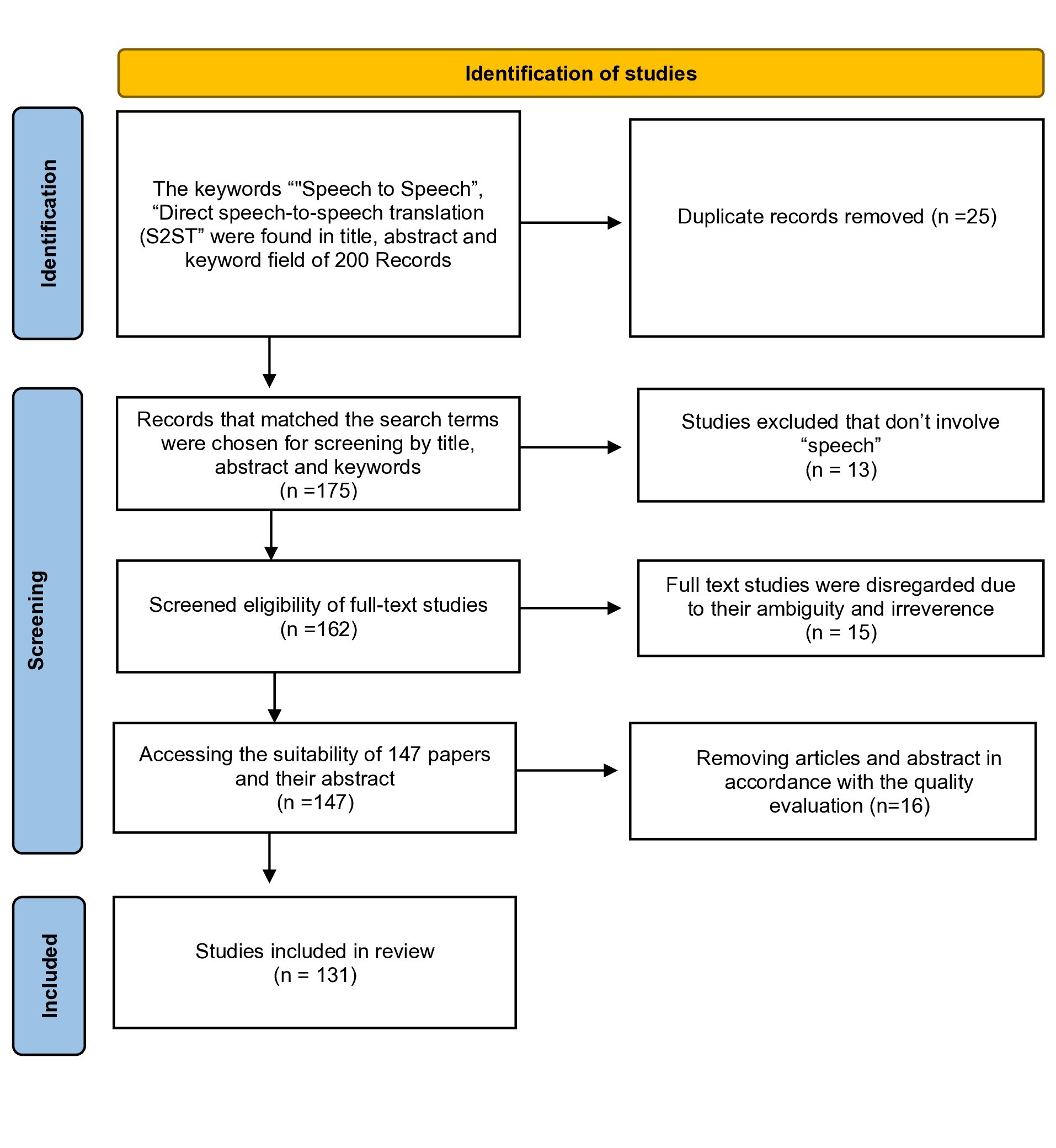} 
    \caption{The adopted PRISMA methodology}
    \label{fig:Prisma_Flow}
\end{figure*}

This review paper is a systematic approach towards the critical review of recent state-of-the-art advancements in direct Speech-to-Speech Translation (S2ST) enabled by machine learning techniques. It includes extensive literature search, clearly articulated inclusion and exclusion criteria, and rigorous assessment protocols ensuring proper integration of relevant research and high-quality research for an effective overview. Relevant articles for this analysis is collected through in-depth searching on many reputed electronic databases, such as IEEE Xplore, ScienceDirect by Elsevier, arXiv, SpringerLink, Google Scholar, and the International Journal of Computer Applications Technology and Research (IJCATR). Our methodology is inspired by \cite{book}. This particular database has relevance as far as providing unique articles based on recent research with topics like machine learning, speech processing, and multilingual learning through conference papers, journals, and preprints. To effectively do systematic review we have followed PRISMA architecture Figure~\ref{fig:Prisma_Flow}.
\\\\
The research approach undertaken in this paper was to apply chosen keywords and Boolean operators, both in general and special aspects of the study domain. Some terms used include "Speech to Speech", "Direct speech-to-speech translation (S2ST)", "Machine learning models for S2ST", "End-to-end neural networks", "Translatotron 2", "Speech translation datasets", "Low-resource language translation", and "Multilingual speech processing". Boolean operators used in this paper were "AND", "OR", and "NOT" to influence the search results so that only relevant sources were retrieved.

\begin{figure*}
    \centering
    \includegraphics[width=\textwidth]{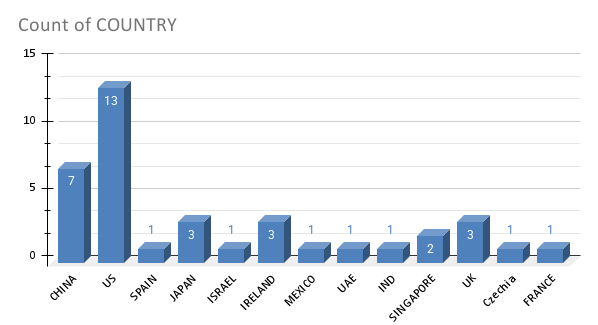} 
    \caption{Various countries along with their counts of works}
    \label{fig:countcountry}
\end{figure*}
The inclusion and exclusion criteria were established in such a manner that only those studies could be considered that matched the focus of the review on direct S2ST. The criteria for inclusion included:
\begin{itemize}
    \item Articles published between 2016 and 2024 related to the state-of-the-art work done in machine learning regarding S2ST.
    \item A survey of direct S2ST techniques that avoid intermediate text representations. 
    \item Studies on the application of models like Translatotron 2, UnitY, or other neural network structures.
    \item Datasets to include diverse ones, such as multilingual, low-resource, and synthetic corpora.
    \item Papers on problems of S2ST such as scarcity, latency, or acoustic-linguistic diversity.
\end{itemize}
Exclusion criteria were:
\begin{itemize}
    \item Articles on cascade models regarding S2ST that concern the conversion of intermediate texts. 
    \item Papers that do present the machine translation for text or audio-only tasks without mentioning S2ST at all.
    \item Papers in which the experiments were scanty or those that relied solely on outdated datasets. 
\end{itemize}

The extraction and analysis were performed in stages to cover and ensure the data as much as possible. The process began with selecting article titles and abstracts that meet the review criteria. Articles that passed muster have their text analyzed followed by the extraction of methodologies, datasets, and architecture models, then studies could be summarized based on selected themes, which included model strategy, challenges, and application. Knowing that S2ST datasets and models contain implicit biases, the review focuses on studies of underrepresented languages and acoustic variation. For instance, CoVoST-2 and Fisher Spanish-English were examined in terms of their suitability to low-resource language environments.

The performance analysis methodically used critical metrics, such as BLEU scores, Mean Opinion Scores (MOS) \cite{articleMOS}\cite{wang2024uncertaintyawaremeanopinionscore}, and latency evaluations \cite{latency}. Furthermore, the data augmentation methodology and support for multiple languages by self-supervised learning methods such as HuBERT and Wav2Vec 2.0 are discussed. The methodological approach is systematic analysis and careful choice of relevant research to ensure that the synthesis is complete and enlightening on progress made in direct S2ST using machine learning.
\section{Datasets}

The 27 studies reviewed collectively utilize a diverse range of datasets for direct speech-to-speech translation (S2ST), predominantly focusing on Spanish-English and English-Spanish pairs. Japanese-English and multilingual combinations are less frequently explored. Figure~ \ref{fig:countcountry} shows country wise contributions to the dataset. The most commonly used datasets include Fisher Spanish-English \cite{shi2023enhancingspeechtospeechtranslationmultiple}, \cite{inaguma2023unitytwopassdirectspeechtospeech}, \cite{zhang2020uwspeechspeechspeechtranslation}, \cite{lee2022directspeechtospeechtranslationdiscrete}, \cite{jia2022translatotron2highqualitydirect} ,\cite{li2022textlessdirectspeechtospeechtranslation}, VoxPopuli \cite{zhu2023diffs2utsemanticpreservingdiffusion}, \cite{wei2022jointpretrainingspeechbilingual}, \cite{nguyen2022improvingspeechtospeechtranslationunlabeled}, \cite{popuri2022enhanceddirectspeechtospeechtranslation}, CoVoST-2 \cite{wei2022jointpretrainingspeechbilingual}, \cite{fang2023daspeechdirectedacyclictransformer}, \cite{yan2023espnetstv2multipurposespokenlanguage} \cite{nachmani2024translatotron3speechspeech}, \cite{popuri2022enhanceddirectspeechtospeechtranslation}, \cite{duquenne2022tmodulestranslationmoduleszeroshot}, and CVSS \cite{fang2023daspeechdirectedacyclictransformer}, \cite{inaguma2023unitytwopassdirectspeechtospeech}, \cite{Mingote_2023}, \cite{jia2022cvsscorpusmassivelymultilingual}, \cite{jia2022leveragingunsupervisedweaklysuperviseddata}, \cite{li2022textlessdirectspeechtospeechtranslation}. These datasets vary significantly in scope, ranging from conversational telephone data to synthesized and scripted readings. Notably, there is a strong emphasis on publicly available datasets for standardized benchmarking across studies (Table~\ref{tab:public}). However, proprietary datasets, such as the Unpaired Conversational Dataset \cite{nachmani2024translatotron3speechspeech} and various synthetic corpora \cite{jia2019directspeechtospeechtranslationsequencetosequence}, \cite{Quintana2018ADS}, \cite{song2023styles2stzeroshotstyletransfer},  also play a significant role, particularly in zero-shot or style-transfer tasks.
\normalsize 
\begin{table*}
\centering
\footnotesize
\caption{Publicly available datasets used to develop direct S2ST models}
\resizebox{\textwidth}{!}{%
\begin{tabular}{|p{2.5cm}|p{1.5cm}|p{2cm}|p{1.5cm}|p{4cm}|p{1.5cm}|p{2cm}|p{2.5cm}|}
\hline
\textbf{Name} & \textbf{Ref.} & \textbf{Used in} & \textbf{Year} & \textbf{Language} & \textbf{Gender} & \textbf{Hours of Speech} & \textbf{Syn/Org} \\\hline
VoxPopuli S2S & \cite{wang-etal-2021-voxpopuli} & {\cite{zhu2023diffs2utsemanticpreservingdiffusion}}, {\cite{wei2022jointpretrainingspeechbilingual}}, {\cite{popuri2022enhanceddirectspeechtospeechtranslation}} & 2021 & 23 languages: Bulgarian, Czech, Danish, Dutch, English, Estonian, Finnish, French, German, Greek, Hungarian, Italian, Latvian, Lithuanian, Polish, Portuguese, Romanian, Slovak, Slovenian, Spanish, Swedish, Croatian & Not specified & 400,000 hours (unlabeled); 1,800 hours (transcribed); 17,300 hours (aligned oral interpretations) & Original \\\hline
Fisher Spanish-English & \cite{post-etal-2013-improved} & {\cite{shi2023enhancingspeechtospeechtranslationmultiple}}, {\cite{inaguma2023unitytwopassdirectspeechtospeech}}, {\cite{zhang2020uwspeechspeechspeechtranslation}}, {\cite{dong2022leveragingpseudolabeleddataimprove}}, {\cite{lee2022directspeechtospeechtranslationdiscrete}}, {\cite{jia2022translatotron2highqualitydirect}}, {\cite{li2022textlessdirectspeechtospeechtranslation}} & 2013 & Spanish and English & Not specified & Approximately 38 hours & Original \\\hline
Europarl-ST & \cite{iranzosánchez2020europarlstmultilingualcorpusspeech} & {\cite{wei2022jointpretrainingspeechbilingual}}, {\cite{nguyen2022improvingspeechtospeechtranslationunlabeled}} & 2020 & 6 European languages: English, French, German, Spanish, Italian, Dutch & Not specified & Not specified & Original \\\hline
CVSS & \cite{jia2022introducing} & {\cite{fang2023daspeechdirectedacyclictransformer}}, {\cite{Mingote_2023}}, {\cite{jia2022cvsscorpusmassivelymultilingual}} & 2022 & 21 languages to English & Not specified & (1,872+1,973) Hours & Synthetic \\\hline
CVSS-C & \cite{jia-etal-2022-cvss} & {\cite{huang2023avtranspeechaudiovisualrobustspeechtospeech}}, {\cite{inaguma2023unitytwopassdirectspeechtospeech}}, {\cite{jia2022leveragingunsupervisedweaklysuperviseddata}}, {\cite{li2022textlessdirectspeechtospeechtranslation}} & 2022 & 21 languages to English & Single high-quality canonical voice & 1,872 Hours & Synthetic \\\hline
LRS3-T & \cite{afouras2018lrs3tedlargescaledatasetvisual} & {\cite{huang2023avtranspeechaudiovisualrobustspeechtospeech}} & 2018 & English & Not specified & Over 400 hours & Original \\\hline
Multi-Domain English-Spanish & \cite{chen2021gigaspeechevolvingmultidomainasr} & {\cite{inaguma2023unitytwopassdirectspeechtospeech}} & 2021 & English & Not specified & 10,000 hours (transcribed); 40,000 hours (total audio) & Original \\\hline
MuST-C(MuST-C v1 \& MuST-C-v2) & \cite{di-gangi-etal-2019-must} & {\cite{yan2023espnetstv2multipurposespokenlanguage}}, {\cite{duquenne2022tmodulestranslationmoduleszeroshot}} & 2019 & English to 8 languages: Dutch, French, German, Italian, Portuguese, Romanian, Russian, Spanish & Not specified & At least 385 hours per language pair & Original \\\hline
BTEC Corpus & \cite{BTEC} & {\cite{tjandra2019speechtospeechtranslationuntranscribedunknown}} & 2003 & Multiple languages & Not specified & Approximately 588,000 utterance-style expressions & Original \\\hline
HCRC Map Task & \cite{anderson1991hcrc} & {\cite{akira16_interspeech}}, {\cite{akira17_interspeech}}, {\cite{hayakawa-etal-2018-speech}} & 1991 & English & Not specified & 128 dialogues & Original \\\hline
ILMT-s2s Corpora & \cite{hayakawa-etal-2016-ilmt} & {\cite{akira16_interspeech}}, {\cite{akira17_interspeech}}, {\cite{hayakawa-etal-2018-speech}} & 2016 & English, Italian & Not specified & Not specified & Original \\\hline
CoVoST 2 & \cite{wang2020covost2massivelymultilingual} & {\cite{nguyen2022improvingspeechtospeechtranslationunlabeled}}, {\cite{nachmani2024translatotron3speechspeech}}, {\cite{popuri2022enhanceddirectspeechtospeechtranslation}}, {\cite{duquenne2022tmodulestranslationmoduleszeroshot}}, {\cite{jia2022translatotron2highqualitydirect}}, {\cite{song2023styles2stzeroshotstyletransfer}} & 2020 & 21 languages to English; English to 15 languages & Not specified & Over 2,200 hours & Original \\\hline
mTEDx & \cite{salesky2021mtedx} & {\cite{nguyen2022improvingspeechtospeechtranslationunlabeled}} & 2021 & Spanish, French, Portuguese, Italian, Russian, Greek, Arabic, German & Gender annotations available for French talks & Varies by language; e.g., Spanish: 35 hours, French: 34 hours, Portuguese: 29 hours & Original \\\hline
MLS & \cite{Pratap_2020} & {\cite{nguyen2022improvingspeechtospeechtranslationunlabeled}}, {\cite{popuri2022enhanceddirectspeechtospeechtranslation}} & 2020 & English, German, Dutch, Spanish, French, Italian, Portuguese, Polish & Not specified & English: $\sim$44,500 hours; Other languages combined: $\sim$6,000 hours & Original \\\hline
CommonVoice & \cite{commonvoice2023} & {\cite{nguyen2022improvingspeechtospeechtranslationunlabeled}}, {\cite{nachmani2024translatotron3speechspeech}}, {\cite{popuri2022enhanceddirectspeechtospeechtranslation}} & 2021 & Not specified & Includes demographic metadata & Not specified & Original \\\hline
LibriSpeech & \cite{7178964} & {\cite{nguyen2022improvingspeechtospeechtranslationunlabeled}}, {\cite{popuri2022enhanceddirectspeechtospeechtranslation}} & 2015 & English & Not specified & Approximately 1,000 hours & Original \\\hline
TEDLIUM & \cite{Hernandez_2018} & {\cite{nguyen2022improvingspeechtospeechtranslationunlabeled}}, {\cite{popuri2022enhanceddirectspeechtospeechtranslation}} & 2012 & English & Not specified & Ranges from 118 to 452 hours & Original \\\hline
TedEn2Zh & \cite{liu19d_interspeech} & {\cite{dong2022leveragingpseudolabeleddataimprove}} & 2022 & English and Chinese & Not specified & Not specified & Original \\\hline
CCMatrix, CCNet & \cite{wenzek2019ccnetextractinghighquality} & {\cite{duquenne2022tmodulestranslationmoduleszeroshot}} & 2020 & Multiple language pairs & Not applicable & Not applicable & Synthetic \\\hline
\end{tabular}%
}
\label{tab:public}
\end{table*}

\begin{table*}
\centering
\caption{Proprietary datasets used to develop direct S2ST models} 
\resizebox{\textwidth}{!}{%
\begin{tabular}{|p{4cm}|p{2cm}|p{1cm}|p{3cm}|p{2cm}|p{4cm}|p{2cm}|}
\hline
\textbf{Name} &
  \textbf{Reference} &
  \textbf{Year} &
  \textbf{Language} &
  \textbf{Gender} &
  \textbf{Hours of Speech} &
  \textbf{Syn/Org} \\ \hline
Direct speech-to-speech translation with a sequence-to-sequence model &
  \cite{jia2019directspeechtospeechtranslationsequencetosequence} &
  2019 &
  Spanish-to-English &
  Not Specified &
  1.4k hours of Spanish source speech and 619 hours of English TTS-synthesized target speech &
  Original \\ \hline
A Direct Speech-to-Speech Neural Network Methodology for Spanish-English Translation &
  \cite{Quintana2018ADS} &
  2018 &
  Spanish-to-English &
  5 Male, 1 Female &
  25 hours training data and 15 hours re-training data &
  Original \\ \hline
Analysis of Layer-Wise Training in Direct Speech to Speech Translation Using \{BI-LSTM\} &
  \cite{DBLP:conf/ococosda/AryaAMP22} &
  2022 &
  English-to-Hindi &
  Not Specified &
  7.76 hours &
  Original \\ \hline
\end{tabular}%
}
\label{tab:Proprietary}
\end{table*}

\subsection{Characteristics}
\subsubsection{Public vs. Proprietary}
Most of the datasets used are publicly available, supporting reproducibility and enabling consistent evaluation in S2ST research. Publicly available datasets include Fisher Spanish-English, CVSS, VoxPopuli, CoVoST-2, and Must-C \cite{inaguma2023unitytwopassdirectspeechtospeech}, \cite{popuri2022enhanceddirectspeechtospeechtranslation}, \cite{lee2022directspeechtospeechtranslationdiscrete}, \cite{jia2022translatotron2highqualitydirect}. Proprietary datasets, such as the Unpaired Conversational Dataset (UC) \cite{nachmani2024translatotron3speechspeech}, ILMT-s2s corpus \cite{akira17_interspeech}, and various synthesized data collections \cite{jia2019directspeechtospeechtranslationsequencetosequence}, \cite{DBLP:conf/ococosda/AryaAMP22}, \cite{song2023styles2stzeroshotstyletransfer}, are typically used for specific purposes, such as fine-tuning or unsupervised S2ST applications, allowing for controlled or specialized training data. While reliance on proprietary datasets limits reproducibility, such data often provides valuable insights for model development. Proprietary dataset have been studied in Table~\ref{tab:Proprietary}.

\subsubsection{Language Coverage}
The representation of languages in the datasets is often biased and skewed, with a dominant focus on Spanish-English and English-French pairs \cite{zhu2023diffs2utsemanticpreservingdiffusion}, \cite{wei2022jointpretrainingspeechbilingual}, \cite{zhang2020uwspeechspeechspeechtranslation}, \cite{lee2022directspeechtospeechtranslationdiscrete}, \cite{jia2022translatotron2highqualitydirect}.
While some datasets, like CVSS and CoVoST-2, make an effort to include low-resource languages alongside high-resource pairs \cite{fang2023daspeechdirectedacyclictransformer}, \cite{inaguma2023unitytwopassdirectspeechtospeech}, \cite{Mingote_2023}, such representations remain limited. This imbalance is particularly evident in low-resource language settings, where the available data is often less comprehensive, impacting the ability of models to generalize effectively across diverse languages.

\subsubsection{Size and Scale}
The size of datasets varies significantly, ranging from smaller collections, such as the HCRC Map Task with only a few hours of speech \cite{akira16_interspeech}, \cite{Quintana2018ADS}, to large multilingual datasets like CVSS, which contains up to 20,000 hours \cite{Mingote_2023}, \cite{jia2022cvsscorpusmassivelymultilingual}, \cite{jia2022leveragingunsupervisedweaklysuperviseddata}, and VoxPopuli, with 100,000 hours \cite{zhu2023diffs2utsemanticpreservingdiffusion}, \cite{wei2022jointpretrainingspeechbilingual}, \cite{nguyen2022improvingspeechtospeechtranslationunlabeled}. While these large datasets provide ample data for pretraining and benchmarking, smaller datasets focusing on conversational data, such as the Fisher Spanish-English dataset with around 170 hours \cite{inaguma2023unitytwopassdirectspeechtospeech}, \cite{zhang2020uwspeechspeechspeechtranslation}, \cite{jia2022translatotron2highqualitydirect}, are equally valuable in real-world application scenarios.

This variation in dataset size highlights the balance between broad multilingual coverage and specialized, context-rich data.

\subsection{Annotations and Preprocessing}
Annotations in most datasets typically include transcriptions, translations, or both. For instance, datasets like Fisher Spanish-English provide detailed conversational transcriptions \cite{shi2023enhancingspeechtospeechtranslationmultiple}, \cite{lee2022directspeechtospeechtranslationdiscrete}. In cases where bilingual S2ST data is not available, synthetic translations are generated using TTS systems to produce target speech, as seen in datasets like CVSS-C \cite{fang2023daspeechdirectedacyclictransformer}, \cite{jia2022leveragingunsupervisedweaklysuperviseddata} and Fisher \cite{lee2022directspeechtospeechtranslationdiscrete}, \cite{li2022textlessdirectspeechtospeechtranslation}. Preprocessing operations often involve steps such as noise reduction, normalization, tokenization, and the use of models like HuBERT or wav2vec for feature extraction \cite{wei2022jointpretrainingspeechbilingual}, \cite{Mingote_2023}, \cite{jia2022cvsscorpusmassivelymultilingual}. Additionally, some datasets incorporate more advanced augmentation techniques, including SpecAugment and Effects-aug, to simulate various environmental conditions and further enhance the robustness of the data \cite{nguyen2022improvingspeechtospeechtranslationunlabeled}, \cite{jia2022translatotron2highqualitydirect}.

\subsection{Data Sources and Modalities}
\subsubsection{Sources}
Data sources for S2ST datasets vary significantly in the type of environment they represent. Real-world contexts are captured in datasets like Fisher Spanish-English, which includes telephone conversations \cite{shi2023enhancingspeechtospeechtranslationmultiple}, \cite{zhang2020uwspeechspeechspeechtranslation}, \cite{jia2022translatotron2highqualitydirect}, while parliamentary proceedings form the basis of datasets such as VoxPopuli \cite{zhu2023diffs2utsemanticpreservingdiffusion}, \cite{wei2022jointpretrainingspeechbilingual}, \cite{nguyen2022improvingspeechtospeechtranslationunlabeled}. In contrast, synthesized corpora, using TTS-based translations, are featured in datasets like CVSS and proprietary conversational datasets \cite{fang2023daspeechdirectedacyclictransformer}, \cite{jia2019directspeechtospeechtranslationsequencetosequence}, \cite{jia2022leveragingunsupervisedweaklysuperviseddata}. Large datasets, such as Europarl-ST \cite{wei2022jointpretrainingspeechbilingual}, \cite{nguyen2022improvingspeechtospeechtranslationunlabeled}, \cite{popuri2022enhanceddirectspeechtospeechtranslation}, primarily consist of scripted content, including TED talks and parliamentary speeches. Although merging natural and synthetic data has proven instrumental in expanding S2ST research, especially for low-resource languages, this approach can introduce limitations in capturing the spontaneous variability characteristic of conversational speech \cite{li2022textlessdirectspeechtospeechtranslation}. Publicly available dataset have been critically analyzed in Table~\ref{tab:public}.

\subsubsection{Modalities}
Most datasets used in S2ST research are audio-only, but a few, such as LRS3-T\cite{huang2023avtranspeechaudiovisualrobustspeechtospeech} and HCRC Map Task \cite{akira16_interspeech}, \cite{akira17_interspeech}, incorporate multimodal features, including visual data like lip movements. This multimodal aspect is particularly significant for models designed to address audio-visual S2ST, as it enhances the alignment between audio and visual cues. By incorporating visual features, these data.
sets contribute to improving the naturalness and contextual accuracy of translations, particularly in conversational scenarios where non-verbal cues play a vital role. 

\subsection{Ethical Consideration}
Many datasets create ethical concerns, specifically on the issues of consent and privacy \cite{articleConcern}. More particularly, where datasets contain human participants, concerns become more stringent, especially for public collections. Such cases include Common Voice and VoxPopuli \cite{zhu2023diffs2utsemanticpreservingdiffusion}, \cite{Mingote_2023}, \cite{jia2022leveragingunsupervisedweaklysuperviseddata}. Synthetic datasets have fewer debates in matters of ethics, although certain studies about privacy, even for synthetic data, are being recommended for implementation \cite{fang2023daspeechdirectedacyclictransformer}, \cite{jia2022leveragingunsupervisedweaklysuperviseddata}. Datasets based on crowd-sourced recordings, as in the case of CVSS and Common Voice, explicitly rely on contributors' opt-in consent to ensure voluntary participation. Others, like Europarl-ST, base their data on parliamentary sources, on which consent is assumed due to the public nature of those proceedings \cite{wei2022jointpretrainingspeechbilingual}. Despite these efforts, biases in demographic representation and language usage still exist and limit the generalization of models across disparate user groups and their applicability to real-world scenarios.
\begin{figure*}
    \centering
    \includegraphics[width=1.0\textwidth]{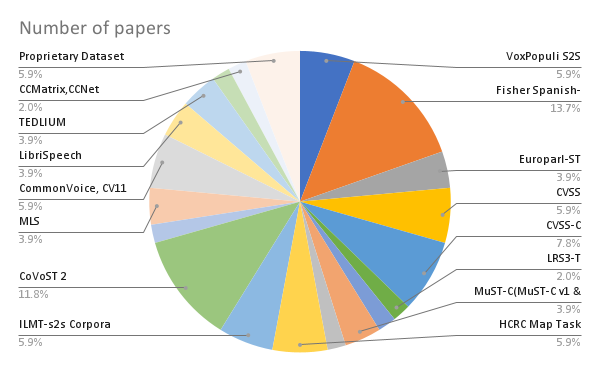}
    \caption{ Visualization of number of papers with its datasets}
    \label{fig:Number_paper}
\end{figure*}
\subsection{Limitations and Benchmarks}
\subsubsection{Common Limitations}
Several limitations in the S2ST datasets are presented \cite{ijcat}: linguistic biases, size constraints, and an overreliance on artificial, formal, or synthesized dialogue rather than natural conversational exchanges. The overrepresentation of specific language pairs, such as Spanish-English or English-French, even further restricts the applicability of cross-lingual models to underrepresented languages, limiting their utility in diverse linguistic contexts \cite{fang2023daspeechdirectedacyclictransformer}, \cite{nguyen2022improvingspeechtospeechtranslationunlabeled}, \cite{popuri2022enhanceddirectspeechtospeechtranslation}. Furthermore, making use of the synthetic data carries overfitting possibilities to the less spontaneous speech, which reduces the effectiveness of the models in real-world conversations. The great drawback of most datasets is that they do not contain multimodal cues, meaning that the models cannot utilize any visual information that is very important in situations where non-verbal communication holds much weight.

\subsubsection{Benchmarking Datasets}
Due to their availability and wide linguistic coverage, frequent benchmarking datasets such as Fisher Spanish-English, VoxPopuli, and CVSS have become the standard ones in S2ST research \cite{zhu2023diffs2utsemanticpreservingdiffusion}, \cite{inaguma2023unitytwopassdirectspeechtospeech}, \cite{zhang2020uwspeechspeechspeechtranslation}, \cite{lee2022directspeechtospeechtranslationdiscrete}. 
These datasets, see figure \ref{fig:Number_paper} allow for comparative analyses across studies and critically contribute toward validation of S2ST model robustness while boosting improvements in BLEU scores and ASR metrics \cite{Sasindran_2024}\cite{roy2021semanticwerunifiedmetricevaluation}. Datasets like CoVoST-2 and Must-C are particularly precious for benchmarking cross-lingual performance, offering critical insights for low-resource translation research and helping advance the field's understanding of multilingual model effectiveness \cite{wei2022jointpretrainingspeechbilingual}, \cite{duquenne2022tmodulestranslationmoduleszeroshot}.

\section{Feature Extraction}
\subsection{Speech Discrete Unit Extraction}
Speech discrete unit extraction is a process that converts continuous speech signals into discrete, and individual units that are useful for the many speech processing tasks performed in recognition, translation, or synthesis. This method effectively simplifies the handling of complex speech data since it separates it into more manageable components \cite{chang2024interspeech2024challengespeech} \cite{10446418} \cite{poncelet2024efficientextractionnoiserobustdiscrete}.
\\\\
HuBERT stands for Hidden Unit BERT model \cite{hsu2021hubertselfsupervisedspeechrepresentation}. This model extracts the discrete units of speech, and in this case, application of K-Means clustering \cite{Jin2010} \cite{5453745} to model outputs gives it self-supervised learning. This tends to differentiate the linguistic features from other acoustic features that are based on speaker's identity or prosody of speech. These discrete units become targets for training S2ST while it enhances its learning. Crucially, this comes with a pre-trained 1,000 K-Means multilingual HuBERT model. A duration predictor recovers the sequence length during synthesis. This lets the improvements to performance metrics like BLEU scores come in by huge margins. Indeed, it has shown strong linguistic normalizing and improved translation quality even with considerable variability in speech signals besides inconsistencies on top of TTS systems \cite{hasanabadi2023overviewtexttospeechsystemsmedia} from tests in datasets like Fisher Spanish-English and LJSpeech \cite{shi2023enhancingspeechtospeechtranslationmultiple}. This masking of audio segments, encoding them into intermediate representations, and K-Means clustering technique aligns these discrete units with the text units in bilingual datasets with a gain of around 3 points BLEU over spectrogram-based methods. Robustness is enhanced using masking and a 16 kHz sampling rate, which alleviates alignment challenges during pre-training with bilingual data \cite{wei2022jointpretrainingspeechbilingual}.
The integration of HuBERT with K-Means clustering allows for self-supervised pretraining using unlabeled speech data, which produces discrete units that capture both linguistic and prosodic characteristics. This approach simplifies complex unit sequences, thus improving efficient encoding and decoding processes, which yields an improvement of up to 2 points in the BLEU score for Speech-to-Unit Translation (S2UT) tasks. The preprocessing stage has involved standardized audio normalization, and this model has been tested using CoVoST-2 and mTEDx among others. Evaluation in those datasets has demonstrated capabilities that would work in low-resource settings. Nonetheless, limited availability of labeled data to fine-tune somewhat ameliorates the problem because the technique is self-supervised \cite{nguyen2022improvingspeechtospeechtranslationunlabeled}, \cite{popuri2022enhanceddirectspeechtospeechtranslation}.
In addition to that, HuBERT units normalize speech by removing speaker identity and prosody, in order to provide more consistency for translations across S2ST systems. Added into vocoders, such units are shown to achieve highly competitive BLEU scores for zero-shot tasks, solving the variability for speech input and exhibiting powerful zero-shot transfer capability on datasets like CoVoST-2 and MLS \cite{duquenne2022tmodulestranslationmoduleszeroshot}. Besides that, discrete speech representations, such as Wav2Vec2, HuBERT, and WavLM, which are included in the S2ST pipeline replace the traditional acoustic features; they effectively incorporate both semantic and phonetic information. The ESPnet-ST-v2 makes use of such representations with the help of tailored frame rates and frequency resolutions. The ablation study indicates the better performance of WavLM in ASR-BLEU scores and the optimal effectiveness at the unit granularity level \cite{yan2023espnetstv2multipurposespokenlanguage}.
\\\\
VQ-VAE\cite{oord2018neuraldiscreterepresentationlearning} generates discrete speech representations optimized for spectrogram reconstruction, which are used by linguistic decoders in S2ST systems. Speech is mapped to a latent space and then approximated with a codebook to guide autoregressive target speech prediction. Experimental settings, including a codebook size of 512, demonstrated a +18.5 BLEU improvement on the Fisher Spanish-English dataset, while hyperparameter tuning helps mitigate potential information loss. In contrast, the Random Quantizer \cite{chiu2022selfsupervisedlearningrandomprojectionquantizer} \cite{randomQuantizationarticle}, a handcrafted approach inspired by BEST-RQ \cite{whetten2024openimplementationstudybestrq}, applies fixed projection matrices and codebook quantization to speech. Although simple, it supports S2ST pipelines without textual supervision but lags in BLEU scores compared to learned quantizers, highlighting the need for optimized techniques in linguistic encoding . Learned quantizers, such as Linear and Transformer-Based Quantizers, further improve discrete speech representation. Linear quantizers optimize projection matrices, while transformer-based quantizers use non-causal transformer stacks to better encode linguistic structures. Their effectiveness is evident in tasks such as Textless Translatotron, with BLEU scores for transformer-based quantizers being the highest among discrete techniques. Success depends on carefully balancing codebook size and stride, and these quantizers perform even better in high-resource settings \cite{li2022textlessdirectspeechtospeechtranslation}. 

\subsection{Mel-Spectrogram Based Feature Extraction}
Mel-Spectrogram based feature extraction: The very first step is to transform the audio signal into a visual representation of the Mel-Spectrogram. These steps include pre-emphasis filtering, framing, windowing, and the Short-Time Fourier Transform (STFT) to obtain the frequency spectrum. Then, the power spectrum is passed through a Mel-scale filter bank to emphasize frequencies similar to human hearing, followed by logarithmic compression. Optional application of the Discrete Cosine Transform (DCT) gives the Mel-Frequency Cepstral Coefficients (MFCCs), which are extremely important in applications like speech recognition and audio analysis.\cite{astuti2022melweighted}\cite{Zhang2019AudioRecognition}
\\\\
Log-mel spectrograms are a time-frequency representation of audio on the Mel scale with logarithmic scaling to match human hearing, preserving critical acoustic information while reducing dimensionality. They are derived by applying Mel filter banks to the frequency spectrum of the audio signal, followed by taking the logarithm of the resultant values \cite{Boulal2024}. \cite{jia2019directspeechtospeechtranslationsequencetosequence} utilized 80-channel Mel spectrograms with adjacent frames stacked to form input features, applying a reduction factor of 2 and enhancing alignment learning by predicting two frames per decoding step. This was combined with data augmentation using noise and reverberation to improve feature robustness, making log-mel spectrograms a primary input for translation and synthesis tasks, despite alignment challenges between source and target spectrograms. Multitask training with an auxiliary phoneme recognition task and corpora like Fisher Spanish-English demonstrated feasibility for direct S2ST with competitive BLEU scores. Similarly, mel-spectrograms, which encode phonetic and prosodic information, are widely employed in S2ST pipelines due to their perceptual frequency representation. DASpeech \cite{fang2023daspeechdirectedacyclictransformer} used an 80-dimensional Mel filterbank for feature extraction, applying cepstral normalization and bridging linguistic and acoustic domains via a FastSpeech 2-based acoustic decoder. XL-VAE \cite{zhang2020uwspeechspeechspeechtranslation} encoded speech as discrete tokens using mel-spectrograms, enhancing phonetic accuracy through cross-lingual transfer, significantly increasing BLEU scores on multilingual datasets like Fisher Spanish-English. ESPnet-ST-v2 \cite{yan2023espnetstv2multipurposespokenlanguage} extracted mel-spectrograms through STFT with a 25 ms window and a 10 ms frame shift, inputting these features to vocoders like HiFi-GAN \cite{kong2020hifigangenerativeadversarialnetworks} for high-quality synthesis while addressing noise sensitivity. Mel-filterbanks further focus on perceptually salient frequencies by projecting audio signals onto the Mel scale. AV-TranSpeech \cite{huang2023avtranspeechaudiovisualrobustspeechtospeech} used 80-dimensional mel-filterbanks to align audio inputs with visual features, improving translation robustness under noisy conditions with training on the LRS3-T dataset. Translatotron \cite{lee2022directspeechtospeechtranslationdiscrete} also employed mel-filterbanks for acoustic representation, although challenges in separating language content from speaker characteristics and prosody led to lower BLEU scores. Across studies, mel-filterbanks and spectrograms demonstrate versatility in capturing acoustic features such as pitch, energy, and duration, enabling effective translation and synthesis in S2ST systems while highlighting the ongoing need for advancements in feature robustness and representation quality.

\subsection{MFCC and Spectrogram Based Features}

Mel-Frequency Cepstral Coefficients (MFCCs) and spectrogram-based features are fundamental in speech signal processing, providing critical linguistic and acoustic representations. MFCCs, derived through framing, windowing, and discrete cosine transform, capture sound characteristics on a perceptual scale tailored to human hearing \cite{mfcc2015}. They have been instrumental in self-supervised learning models like HuBERT \cite{lee2022directspeechtospeechtranslationdiscrete}, where they initiated the representation learning pipeline by mapping masked audio segments to discrete labels, facilitating training refinements that improved translation and synthesis outcomes. MFCCs were also employed as primary features in \cite{tjandra2019speechtospeechtranslationuntranscribedunknown}, generating 39-dimensional feature vectors (including $\Delta$ and $\Delta^2$ derivatives) to train sequence-to-sequence translation models and invert codebooks for target speech reconstruction, effectively addressing length mismatches. Experiments on the BTEC corpus yielded notable BLEU and METEOR scores \cite{articleMeteorScores} for French-English and Japanese-English translations, underscoring MFCCs’ utility in handling untranscribed languages.
\\\\
Spectrogram-based features, generated using short-time Fourier transform (STFT) \cite{articleStft}, capture detailed time-frequency information vital for transcription, translation, and synthesis tasks \cite{kent2002acoustic}. Linear magnitude spectrograms, employed by \cite{tjandra2019speechtospeechtranslationuntranscribedunknown}, supported high-fidelity waveform reconstruction during inference, though Griffin-Lim \cite{Griffin-Lim} phase reconstruction introduced slight naturalness degradation, suggesting the need for neural vocoders. In the ILMT-S2S system \cite{akira16_interspeech}, spectrogram-based features extracted via STFT enabled effective transcription before translation and synthesis, though challenges like ASR errors and latency affected dialogue flow. Enhanced feature robustness was achieved using SpecAugment, which incorporated time warping and frequency masking to reduce overfitting during training \cite{nachmani2024translatotron3speechspeech}. Masked autoencoder mechanisms processed spectrograms as input for encoder operations, improving naturalness and intelligibility, with MOS gains in English-to-Spanish and Spanish-to-English synthesis. Despite their robustness, spectrogram-based features, derived from datasets like UC and Common Voice 11, were computationally intensive and occasionally lost speaker-specific details. Preprocessing, including normalization, ensured aligned speech-text pairs for effective training across systems.

\subsection{Self-Supervised Learning (SSL) Speech Models}

Wav2Vec 2.0 is a self-supervised learning-based approach capable of extracting context-rich representations directly from raw audio, making it highly suitable for downstream tasks such as automatic speech recognition (ASR) and direct speech-to-speech translation (S2ST). The model processes raw audio by segmenting it into frames, encoding them with a CNN \cite{oshea2015introductionconvolutionalneuralnetworks}, and modeling their context using a transformer. A key advantage of Wav2Vec 2.0 is its ability to learn representations directly from data without relying on handcrafted features, significantly enhancing generalization and performance, particularly in low-resource settings. Pre-training is typically performed on large multilingual datasets, a computationally expensive process mitigated by the availability of pre-trained checkpoints, which improves ASR accuracy and BLEU scores for S2ST tasks. Research studies have demonstrated various applications of Wav2Vec 2.0, such as the UnitY S2ST system utilizing Wav2Vec 2.0 pre-trained on datasets like Fisher and CVSS-C to achieve substantial gains in BLEU scores and ASR performance by extracting robust audio features \cite{inaguma2023unitytwopassdirectspeechtospeech}. Another study incorporated Wav2Vec 2.0 with a masking mechanism at the input and a Conformer architecture \cite{gulati2020conformerconvolutionaugmentedtransformerspeech} to enhance efficiency and performance, demonstrating significant improvements in BLEU scores when tested on datasets like Libri-light and VoxPopuli \cite{popuri2022enhanceddirectspeechtospeechtranslation}. Similarly, Wav2Vec 2.0 has been employed in a speech-to-unit module to encode raw audio into meaningful features, enabling better semantic translation and improved translation performance on parallel corpora and unseen test sets \cite{song2023styles2stzeroshotstyletransfer}. HuBERT (Hidden Unit BERT), another self-supervised learning technique, tokenizes speech into discrete units via k-means clustering, learning representations directly from raw audio and capturing both linguistic and prosodic information. Its two-stage training process involves unsupervised clustering to create pseudo-labels followed by masked prediction using a Transformer-based model, which reduces output sequence length, enhancing efficiency and enabling applications such as speech synthesis and semantic translation. Studies have highlighted HuBERT's utility in S2ST pipelines, such as extracting discrete acoustic units for a second-pass decoder, simplifying duration modeling, and improving BLEU scores with a decoding speed-up of 2.51× compared to continuous representations \cite{inaguma2023unitytwopassdirectspeechtospeech}. Another study employed HuBERT for speech-to-unit translation, extracting intermediate representations and clustering them to create semantic-rich discrete units, which significantly improved BLEU scores in low-resource settings when evaluated on datasets like CoVoST-2 and Europarl-ST \cite{nguyen2022improvingspeechtospeechtranslationunlabeled}. Additionally, HuBERT's application to multilingual datasets such as WenetSpeech demonstrated superior semantic alignment and prosodic preservation, evidenced by higher units-BLEU scores in English-Chinese speech translation \cite{song2023styles2stzeroshotstyletransfer}. Multilingual HuBERT (mHuBERT) extends the self-supervised framework of HuBERT to accommodate multilingual data, learning continuous speech representations from large-scale, unlabeled corpora in multiple languages, subsequently discretized using k-means clustering. The resulting discrete units are engineered to capture rich phonetic and linguistic information, making them well-suited for direct S2ST tasks. Studies using mHuBERT have demonstrated its effectiveness in boosting translation performance across different languages, with examples such as extracting continuous representations from English, Spanish, and French audio, clustering into 1,000 centroids to form discrete sequences, preserving both semantic and acoustic information, and leading to significant improvements in the BLEU score on the VoxPopuli-S2S dataset \cite{zhu2023diffs2utsemanticpreservingdiffusion}. Another experiment utilized mHuBERT to extract phonetic and word-level features from the higher layers of a model, collapsing consecutive identical clusters for efficiency, resulting in improved accuracy for text-to-speech and S2ST tasks \cite{Mingote_2023}. The w2v-BERT model employs contrastive learning with a masked language model (MLM) objective to generate speech representations that serve as inputs to S2ST systems from the self-supervised framework, processing input speech using a feature encoder and a context network to project it into a latent space where rich contextual representations are learned. Pre-trained on large multilingual datasets, w2v-BERT demonstrates improvements in translation quality and performance, particularly in low-resource scenarios. Experiments have integrated w2v-BERT with Translatotron 2, applying it to high-quality embeddings extracted from multiple languages based on over 429,000 hours of unsupervised speech data, achieving a 7.8 BLEU score across 21 languages \cite{jia2022cvsscorpusmassivelymultilingual}. Another study used w2v-BERT as the speech encoder in the Textless Translatotron system, showing its capability to handle varying levels of resource availability and achieve near text-supervised BLEU scores when tested on CVSS-C and Fisher Spanish-English datasets \cite{li2022textlessdirectspeechtospeechtranslation}. AV-HuBERT integrates both audio and visual modalities to extract features in multimodal S2ST systems, employing a self-supervised framework to derive contextual representations by predicting masked inputs from synchronous audio and visual data. Audio features are extracted using mel-filterbanks, while visual features are obtained through a modified ResNet applied to lip region frames, with modality fusion enhancing system robustness, particularly in noisy environments and low-resource scenarios. In the AV-TranSpeech system, AV-HuBERT improved the average BLEU score by 7.6 compared to audio-only baselines, addressing temporal resolution mismatches between audio and visual streams using a modality adaptor layer, as demonstrated on datasets like LRS3-T \cite{huang2023avtranspeechaudiovisualrobustspeechtospeech}. The fusion of multiple speech SSL features, such as Wav2Vec 2.0, HuBERT, and WavLM, leverages the strengths of each model to build stronger features in S2ST systems by extracting individual features, normalizing them, and combining them through weighted averaging or concatenation. By utilizing these complementary strengths, fused features enhance translation quality and improve generalization across various acoustic conditions, with the ESPnet-ST-v2 achieving significant improvements in BLEU scores. For example, WavLMHuBERT demonstrated superior performance when fused compared to individual features, despite introducing additional computational complexity \cite{yan2023espnetstv2multipurposespokenlanguage}.

\subsection{Custom and Cross-Lingual Feature Adaptations}

Direct S2ST models utilize discrete unit-based representations derived from SSL frameworks, such as HuBERT and WavLM, which are quantized to encode features into discrete tokens for efficient computation. These units are adapted and specialized for specific tasks, such as speech-to-text and text-to-speech translation, through adjustments to quantization parameters and unit-level duration predictors, ensuring well-aligned input/output sequences critical for maintaining translation fidelity. For example, custom adaptations like the UnitY model have demonstrated significant improvements in ASR-BLEU scores, achieving competitive state-of-the-art performance with a score of 32.0, comparable to existing benchmarks \cite{yan2023espnetstv2multipurposespokenlanguage}. Challenges such as maintaining naturalness and stability in synthesized speech are addressed using vocoders and duration modeling techniques, with data preprocessing involving unit extraction, normalization, and mapping to target tokens, primarily using the CVSS-C dataset. Similarly, vector quantization techniques like XL-VAE enhance speech processing by encoding continuous features into discrete tokens, capturing phonetic and semantic content through mel-spectrogram inputs processed by convolution layers and Transformer blocks. XL-VAE facilitates cross-lingual speech recognition by enabling phonetic transfer from written to unwritten languages, achieving notable improvements in translation quality, including a 10-point BLEU score gain over VQ-VAE and 16 points over Direct Translation \cite{zhang2020uwspeechspeechspeechtranslation}. Preprocessing steps, such as up-sampling datasets, complement its performance, as demonstrated using the Fisher Spanish-English dataset. Furthermore, multi-modal learning with mSLAM integrates speech features learned via w2v-BERT and text features from SpanBERT \cite{joshi2020spanbertimprovingpretrainingrepresenting}, enabling the model to operate on both unlabeled and paired data. Trained on extensive datasets such as 429k hours of speech, 6 trillion text tokens from mC4, and 2.4k hours of paired speech-transcript data, mSLAM improves cross-modal alignment, achieving an average BLEU score improvement of +1.3 and a remarkable +7.8 when pre-trained features are utilized during training \cite{jia2022leveragingunsupervisedweaklysuperviseddata}. However, challenges such as overfitting and suboptimal utilization of text features highlight the need for computational resources and alignment tuning. Collectively, these advancements—spanning discrete units, vector quantization, and multi-modal learning—enhance S2ST applications, enabling robust and high-quality translation workflows \cite{yan2023espnetstv2multipurposespokenlanguage}\cite{zhang2020uwspeechspeechspeechtranslation}\cite{jia2022leveragingunsupervisedweaklysuperviseddata}.

\subsection{Applications in Speech Translation and Synthesis}

Speech rate alignment features and spectrogram-based features play critical roles in S2ST systems by facilitating effective feature extraction and system evaluation. Speech rate alignment begins with feature extraction through a TTS system that generates speech at a standard rate of 180 words per minute (wpm). Utterances are segmented using transcription data and manually annotated time boundaries, with Praat employed for precise time alignment. These features are compared to the duration of human utterances versus TTS durations to derive alignment metrics essential for evaluating system-mediated communication \cite{hayakawa-etal-2018-speech}. While this method demonstrates robustness, achieving high Adjusted R² values, such as 0.948 in predicting speech rate variations for multilingual dialogues between English and Portuguese, it faces challenges like the labor-intensive manual annotation process and handling very short or inconsistent TTS outputs. Complementarily, spectrogram-based features are crucial for transforming speech signals into time-frequency representations within the ILMT-S2S system \cite{ilmt-s2s}. Handcrafted feature extraction using STFT with carefully selected parameters preserves linguistic details and eliminates background noise, serving as input to the ASR module to ensure accurate transcriptions before subsequent translation and synthesis \cite{akira16_interspeech}. These features have shown excellent transcription capabilities in task-oriented dialogues, particularly with the ILMT-S2S corpus focusing on English and Portuguese, despite ASR errors and latency in synthesis. User adaptations, such as slower speech and rephrased utterances, slightly constrained natural dialogue within the system compared to human-mediated communication, but preprocessing steps like audio normalization and annotated transcripts revealed performance patterns and error distributions, aiding in refining the system \cite{akira16_interspeech}\cite{hayakawa-etal-2018-speech}.

\subsection{Speaker Embedding Features}
Recent work has employed speaker embedding features, particularly those derived from ECAPA-TDNN models \cite{Desplanques_2020}, to improve both style retention and speaker adaptation in S2ST systems. These embeddings capture global style features such as timbre, pitch, and rhythm from input utterances and are used as conditioning factors for the acoustic decoder to establish effective style control. With the model pre-trained on a multilingual dataset of 30,000 speakers from languages like English and Chinese, the embedding model creates a strong speaker space that generalizes well across unseen speakers \cite{song2023styles2stzeroshotstyletransfer}. This pretraining helps avoid mismatched speaker spaces between languages, enhancing the zero-shot capabilities of the system. The application of such embeddings was found to significantly increase the SMOS and SECS scores in the ILMT-S2S system, indicating better style adaptation and speaker consistency even in unseen scenarios. These features ensure that synthesized speech outputs are natural and individualistic \cite{li2022textlessdirectspeechtospeechtranslation}.  
The integration of state-of-the-art feature extraction techniques—from speech rate alignment and spectrogram-based inputs to speaker embeddings—forms the backbone of modern S2ST systems, addressing several challenges in naturalness, accuracy, and user adaptation across multilingual settings.

\section{Major ML Models}

\subsection{Sequence-to-Sequence Models} 
The Transformer Encoder-Decoder is a powerful architecture extensively used in NLP, consisting of two main components: an encoder that generates contextual representations of the input sequence using self-attention and a decoder that utilizes both self-attention and encoder-decoder attention to produce the output sequence one element at a time. This architecture, noted for its parallelism, has become standard for tasks like machine translation and text summarization \cite{DBLP:journals/corr/VaswaniSPUJGKP17}. \cite{zhu2023diffs2utsemanticpreservingdiffusion} applied this architecture for encoding source speech and decoding target discrete units, leveraging non-causal attention to enhance processing speed and parallelism. In contrast, mBART (Multilingual BART) combines denoising autoencoders with transformer architecture for multilingual tasks, pre-trained on massive datasets \cite{DBLP:journals/corr/abs-2001-08210}. It has been widely applied, such as in \cite{inaguma2023unitytwopassdirectspeechtospeech}, where it was used to pre-train the text decoder in UnitY for multilingual text representation, and in \cite{nguyen2022improvingspeechtospeechtranslationunlabeled}, where mBART served as a decoder in the S2UT model for translating speech into discrete units. \cite{popuri2022enhanceddirectspeechtospeechtranslation} further modified mBART for improved discrete speech unit predictions in speech translation. CRISS (Cross-lingual Retrieval for Iterative Self-Supervised Training) focuses on cross-lingual sentence representation through unsupervised machine translation and iterative training to align embeddings across languages \cite{DBLP:journals/corr/abs-2006-09526}. \cite{nguyen2022improvingspeechtospeechtranslationunlabeled} utilized CRISS to synthesize source-target text pairs by translating source text of unlabeled target text data, subsequently enhancing the S2ST training dataset for better monolingual text generation. The UnitY model combines a Conformer encoder with an mBART-based decoder for multilingual speech-to-speech translation, transforming source speech into discrete phonetic units and performing a two-pass decoding process \cite{inaguma2023unitytwopassdirectspeechtospeech}
. \cite{yan2023espnetstv2multipurposespokenlanguage} highlights its efficiency for resource-constrained environments and multilingual speech processing, while ESPnet-ST-v2 uses a vocoder to convert UnitY’s discrete speech units into natural target audio speech. Finally, the PnG NAT (Phoneme and Grapheme Non-Autoregressive Transformer) improves efficiency and quality in speech synthesis by incorporating phonemic and graphemic inputs, enabling parallel speech generation for faster processing \cite{DBLP:journals/corr/abs-2103-15060}. \cite{jia2022cvsscorpusmassivelymultilingual} used PnG NAT for synthesizing English translations in the CVSS corpus, preserving speaker identity and improving pronunciation and prosody through phoneme and grapheme inputs.

\subsection{Speech Synthesis Models}
Tacotron 2, FastSpeech 2, and VITS are advanced TTS models, each contributing unique strengths to speech synthesis. Tacotron 2, an encoder-decoder autoregressive model with attention mechanisms, converts text into mel-spectrograms that are synthesized into high-quality, natural-sounding speech using a WaveNet vocoder \cite{shen2018naturalttssynthesisconditioning} \cite{oord2016wavenetgenerativemodelraw}. In \cite{shi2023enhancingspeechtospeechtranslationmultiple}, Tacotron 2 was used alongside non-autoregressive systems like FastSpeech 2 and VITS to produce diverse linguistic and acoustic features, enhancing lexical consistency across various speech styles. FastSpeech 2, a non-autoregressive model, synthesizes speech in parallel by predicting pitch, duration, and energy for each phoneme, offering greater speed and stability \cite{ren2022fastspeech2fasthighquality}. It has been utilized in \cite{shi2023enhancingspeechtospeechtranslationmultiple} to diversify synthesized speech and in \cite{fang2023daspeechdirectedacyclictransformer} and \cite{song2023styles2stzeroshotstyletransfer} for tasks like multilingual TTS and style transfer across languages. VITS, a text-to-waveform model, unifies linguistic and acoustic modeling using variational inference and adversarial training, generating expressive, natural prosody in synthesized speech \cite{kim2021conditionalvariationalautoencoderadversarial}. Its integration with Tacotron 2 and FastSpeech 2 in \cite{shi2023enhancingspeechtospeechtranslationmultiple} highlights its ability to enhance variability and translation quality, achieving robust and consistent speech synthesis across diverse styles.

\subsection{Speech Recognition and Understanding Models}
AV-HuBERT, HuBERT, w2v-BERT, Conformer, and UWSpeech are advanced models designed for speech processing and translation tasks, each addressing specific challenges in the field. AV-HuBERT extends HuBERT by integrating audio and visual features, such as lip movements, to enhance robust speech representations \cite{shi2022learningaudiovisualspeechrepresentation}, as used in \cite{huang2023avtranspeechaudiovisualrobustspeechtospeech} for AV-S2ST and AV-TranSpeech to overcome data scarcity and maintain translation quality in noisy environments. HuBERT, a self-supervised model, learns speech representations through masked prediction and clustering, extracting phonetic and semantic insights for tasks like ASR and S2ST \cite{hsu2021hubertselfsupervisedspeechrepresentation}. It has been applied in \cite{huang2023avtranspeechaudiovisualrobustspeechtospeech} for multimodal inputs, in \cite{Mingote_2023} for TTS without text, and in \cite{nguyen2022improvingspeechtospeechtranslationunlabeled} for processing discrete units in S2UT. w2v-BERT combines Wav2Vec 2.0’s robust acoustic feature extraction with BERT’s contextual language understanding \cite{chung2021w2vbertcombiningcontrastivelearning}, enhancing models like Translatotron 2 \cite{jia2022leveragingunsupervisedweaklysuperviseddata} and Textless Translatotron \cite{li2022textlessdirectspeechtospeechtranslation} for low-resource language processing and efficient feature extraction. The Conformer, merging convolutional layers with transformers, captures both local dependencies and global context, making it suitable for speech recognition and translation tasks \cite{gulati2020conformerconvolutionaugmentedtransformerspeech}. It has been employed in UnitY \cite{inaguma2023unitytwopassdirectspeechtospeech}, ESPnet-ST-v2 \cite{yan2023espnetstv2multipurposespokenlanguage}, and Translatotron 2 \cite{jia2022translatotron2highqualitydirect},\cite{jia2022cvsscorpusmassivelymultilingual} for improved speech contextualization and representation. Lastly, UWSpeech, an unsupervised system for unpaired speech data, enables multilingual and textless speech-to-speech translation by leveraging discrete unit discovery and self-supervised learning \cite{zhang2020uwspeechspeechspeechtranslation}, as demonstrated in \cite{zhang2020uwspeechspeechspeechtranslation} for unwritten languages, bypassing intermediate text or phonetic transcription.

\subsection{Speech-to-Speech Translation Models}
Speech-to-speech translation (S2ST) models have evolved to enable direct source-to-target speech translation without intermediate text representations, leveraging advanced architectures for efficiency and quality. The S2UT model translates source speech into discrete units and reconstructs target speech using a unit-based neural vocoder, bypassing spectrograms and reducing sequence length for more efficient training and inference. \cite{shi2023enhancingspeechtospeechtranslationmultiple} optimized S2UT using HuBERT representations, K-means clustering, and a Conformer-based wav2vec 2.0 encoder with an mBART-based decoder to deliver high-quality target speech. \cite{popuri2022enhanceddirectspeechtospeechtranslation} utilized discrete unit representations for languages without writing forms, employing a speech encoder, unit decoder, and vocoder for speech reconstruction. \cite{lee2022directspeechtospeechtranslationdiscrete} introduced convolutional layers for input downsampling and a cross-entropy loss-optimized decoder, while \cite{nguyen2022improvingspeechtospeechtranslationunlabeled} leveraged an attention-based sequence-to-sequence model with a Wav2Vec 2.0 encoder and unit-mBART decoder for efficient, high-quality translation. Similarly, Translatotron 2 performs direct speech-to-speech translation using Conformer layers for linguistic and acoustic information encapsulation, enhancing translation accuracy and speaker identity preservation. \cite{yan2023espnetstv2multipurposespokenlanguage} demonstrated natural-sounding translations with Mel-spectrogram generation, and \cite{jia2022translatotron2highqualitydirect},\cite{jia2022cvsscorpusmassivelymultilingual},\cite{jia2022leveragingunsupervisedweaklysuperviseddata} employed it as a baseline for improved translation quality and multilingual capabilities, integrating speech encoder, linguistic decoder, and acoustic synthesizer for seamless end-to-end S2ST. StyleS2ST extends this paradigm by transferring linguistic content and speaking style, including tone and prosody, across languages. \cite{song2023styles2stzeroshotstyletransfer} achieved zero-shot style transfer using a speech-to-unit framework with a style adaptor to preserve speaker-specific characteristics like timbre and prosody, enhancing naturalness and style similarity in cross-lingual translation.

\subsection{Generative and Diffusion-Based Models}
Generative and diffusion-based models have been extensively applied in speech-to-speech translation and synthesis due to their ability to produce high-quality and realistic outputs. Diffusion Models, as latent variable generative models, iteratively refine noise signals into target distributions, offering high-quality samples. \cite{zhu2023diffs2utsemanticpreservingdiffusion} adapted Diffusion Models for S2ST by combining continuous Gaussian noise in speech representation with discrete denoising using K-means clustering, preserving semantic content and reducing decoder latency for efficient direct S2ST. Complementing this, DSPGAN (Diffusion-based Speech Prior Generative Adversarial Network) serves as a neural vocoder for synthesizing high-fidelity speech from spectrograms \cite{song2023dspganganbaseduniversalvocoder}. \cite{song2023styles2stzeroshotstyletransfer} demonstrated DSPGAN’s effectiveness in zero-shot style transfer conditions, ensuring naturalness and style retention in multilingual translations. The Vector Quantized Variational Autoencoder (VQ-VAE) further enhances speech translation and synthesis by converting input data into discrete latent representations \cite{oord2018neuraldiscreterepresentationlearning}. \cite{tjandra2019speechtospeechtranslationuntranscribedunknown} used VQ-VAE for unsupervised discovery of speech units, enabling cross-linguistic translation without linguistic labeling, while \cite{li2022textlessdirectspeechtospeechtranslation} introduced a VQ-VAE-based speech quantizer to encode phoneme-like discrete units for textless frameworks and reconstruct translated spectrograms. Additionally, XL-VAE (Cross-Lingual Variational Autoencoder) excels in cross-lingual tasks by learning shared latent representations across languages using variational inference. \cite{zhang2020uwspeechspeechspeechtranslation} extended VQ-VAE to cross-lingual tasks with phonetic labels, improving structured speech tokens for translating unwritten languages efficiently, building on UWSpeech’s framework for unsupervised methods.

\subsection{Multimodal and Multilingual Models}
Multimodal and multilingual models, such as DA-Transformer and mSLAM, have been pivotal in advancing speech translation and cross-lingual comprehension. DA-Transformer, a speech translation model built on a transformer-based architecture, utilizes disentangled acoustic features to separate linguistic and non-linguistic streams, improving the handling of speech variability \cite{huang2022directedacyclictransformernonautoregressive}. \cite{fang2023daspeechdirectedacyclictransformer} employed DA-Transformer for S2ST using a non-autoregressive Transformer with a two-pass network comprising a linguistic decoder for target text and an acoustic decoder for generating mel-spectrograms. This design, leveraging Directed Acyclic Graphs (DAGs) and hidden states over translation paths, achieves faster and more accurate speech-to-speech translation. Similarly, mSLAM, a multilingual speech and language model, is tailored for self-supervised learning with speech and text data across various languages. Pre-trained on unsupervised speech, text, and paired data, it generates uniform representations for tasks like speech-to-text translation and cross-modal comprehension, especially in low-resource language scenarios \cite{bapna2022mslammassivelymultilingualjoint}. In \cite{jia2022leveragingunsupervisedweaklysuperviseddata}, mSLAM's multi-task setup fine-tuned Translatotron 2, leveraging its capability to learn from large, diverse datasets to enhance translation accuracy.

\subsection{Recurrent Neural Networks}
Recurrent Neural Networks (RNNs), particularly Long Short-Term Memory (LSTM) and Bi-LSTM networks, are powerful tools for handling sequential data and capturing temporal dependencies, making them highly effective in speech-to-speech translation and related tasks. LSTM networks, designed to manage long-term dependencies, use memory cells and gate mechanisms to retain important information while filtering out irrelevant data, proving efficient in speech processing and time series analysis. \cite{Quintana2018ADS} demonstrated the use of a four-layer LSTM network with 500 neurons per layer for Spanish-to-English speech-to-speech translation, where spectrograms from Spanish audio are interpreted and vectorized to produce English audio outputs, bypassing intermediate text representation. Bi-LSTM networks further enhance sequential modeling by processing data in both forward and backward directions, capturing past and future context for improved performance in tasks like speech recognition and translation \cite{GRAVES2005602}. \cite{DBLP:conf/ococosda/AryaAMP22} applied Bi-LSTM to direct speech-to-speech translation (DS2ST), where input sequences are mapped between source and target languages without text transcription, demonstrating that adding more Bi-LSTM layers improves performance by leveraging richer contextual information from both speech frames.

\section{ML Techniques}
\subsection{Clustering and Augmentation Techniques}
K-Means Clustering is an unsupervised learning algorithm that partitions data into k clusters based on feature similarity, iteratively assigning data points to the nearest cluster centroid and updating the centroids to minimize intra-cluster variance. It is highly effective in speech and language processing, particularly in tasks such as speech representation and textless translation. In works by \cite{zhu2023diffs2utsemanticpreservingdiffusion}, \cite{shi2023enhancingspeechtospeechtranslationmultiple}, \cite{Mingote_2023}, \cite{popuri2022enhanceddirectspeechtospeechtranslation}, \cite{lee2022directspeechtospeechtranslationdiscrete}, \cite{nguyen2022improvingspeechtospeechtranslationunlabeled}, and \cite{song2023styles2stzeroshotstyletransfer}, K-Means is used with HuBERT and mHuBERT models to derive linguistic intent from speech, supporting textless translation by mapping discrete units with semantic or stylistic attributes. This approach enhances StyleS2ST models and language-agnostic S2ST by separating text-based components of speech, thereby improving overall style-preserving translation and the use of discrete units as intermediate representations in frameworks like StyleS2ST and diffusion models \cite{zhu2023diffs2utsemanticpreservingdiffusion}. Text-Augmentation (Text-Aug) is another technique that improves training sets by creating new versions of existing text data through methods like paraphrasing or synonym replacement. This enhances the model’s robustness and generalization, particularly in natural language processing tasks. \cite{nguyen2022improvingspeechtospeechtranslationunlabeled} applies Text-Augmentation to synthesize speech from raw, unannotated texts for S2ST training, introducing Effects-Augmentation, which further enriches data diversity by adding pitch, pace, and background noise variations. This improves generalization and robustness in low-resource and unsupervised S2ST tasks. Effects-Augmentation (Effects-Aug) simulates real-world conditions by manipulating speech training material with artificial effects such as noise addition or pitch shifts, helping models withstand environmental changes. \cite{nguyen2022improvingspeechtospeechtranslationunlabeled} highlights the use of pitch, speed, and background noise variations in synthetic speech data, improving robustness and generalization for low-resource and unsupervised S2ST environments. SpecAugment is another data augmentation technique used during training that applies temporal distortion, masking, and frequency warping to spectrograms, primarily enhancing the robustness of speech recognition models \cite{Park_2019} \cite{nachmani2024translatotron3speechspeech} uses SpecAugment during the masked autoencoder training phase to improve the generalization of Translatotron 3, enabling it to handle multilingual tasks. \cite{lee2022directspeechtospeechtranslationdiscrete} also applies SpecAugment on source speech data to improve variability and robustness to changes in pitch or speed, preventing overfitting and enhancing translation quality. Similarly, \cite{jia2022translatotron2highqualitydirect} uses SpecAugment to mask parts of the spectrogram, improving resilience to real-world speech variations and enhancing model performance. ConcatAug combines speech segments from different samples to enhance training data, increasing the input space and improving the performance of speech-related tasks. \cite{jia2022translatotron2highqualitydirect} applies ConcatAug to augment data for handling speaker turns in multi-speaker audio inputs by concatenating audio samples from different speakers, which benefits multilingual multi-speaker conversation setups without explicit segmentation of speakers.

\subsection{Regularization and Optimization}
Dropout Regularization is a technique that randomly deactivates neurons during training to mitigate overfitting and enhance generalization. It is widely used in neural networks across various applications, including speech and language processing. \cite{Quintana2018ADS} applied Dropout Regularization to all LSTM layers with a dropout rate of 0.3 to prevent overfitting. This technique randomly resets 30\% of synaptic weights during training, ensuring that the model generalizes well on variations in input data, which helps maintain performance when encountering unseen or new audio samples. R-Drop Regularization introduces randomness during training by computing the loss between predictions from multiple stochastic forward passes of the same input. This combats overfitting and ensures consistent outputs over variations of the input. \cite{inaguma2023unitytwopassdirectspeechtospeech} employed R-Drop Regularization to stabilize the two-pass decoding framework during training, enhancing generalization for discrete symbol forecasting with uniform dropout on successive forward passes. Backpropagation Through Time (BPTT) is an extension of backpropagation used for training RNNs, where the network is unrolled over time. It computes gradients across each time step, optimizing dependencies over time and ensuring effective processing of sequential data \cite{58337}. \cite{DBLP:conf/ococosda/AryaAMP22} used BPTT to train a Bi-LSTM for speech-to-speech translation, capturing temporal dependencies that allowed the model to maintain linguistic coherence and map source and target speech sequences more accurately. This method enables the model to fully exploit speech sequences without losing contextual information across layers.

\subsection{Learning Paradigms}
Speech Self-Supervised Learning (SSL) leverages unannotated speech data to derive robust representations by predicting masked or modified input speech. SSL models are often used as pre-trained encoders for low-resource speech-related tasks. \cite{yan2023espnetstv2multipurposespokenlanguage} demonstrated the use of HuBERT and WavLM SSC models added to ESPnet-ST-v2 to complement input speech representations, enabling the extraction of features without the need for large quantities of labeled data. This approach enhances robustness in low-resource settings and captures the subtleties in speech for better translations. Multitask Learning is a training paradigm where a model is optimized for multiple tasks simultaneously, exploiting task-specific features between tasks for better performance and efficiency \cite{caruana1997multitask}. \cite{jia2019directspeechtospeechtranslationsequencetosequence} utilized multitask learning in the phoneme sequence prediction task, regularizing the attention learning and improving translation quality, especially with spontaneous speech. \cite{lee2022directspeechtospeechtranslationdiscrete} used multitask learning to reinforce the training of the S2UT model with auxiliary tasks like phoneme or character prediction, stabilizing the training process and enhancing linguistic feature capture. \cite{jia2022translatotron2highqualitydirect} employed multitask learning with an auxiliary phoneme prediction task to improve the alignment of source and target speech, enhancing translation quality. A Multi-Task Framework trains one model on multiple related tasks simultaneously, allowing information sharing between tasks to improve performance. It is widely used in speech-to-speech and text-to-speech models, where common feature learning across different data types is beneficial. \cite{shi2023enhancingspeechtospeechtranslationmultiple} used a Multi-Task Framework to enhance the S2ST model with different units in various TTS systems, introducing a special token for each target speech sample to indicate which TTS system it belongs to. This approach improved BLEU scores when training with multiple TTS-generated targets. Mixed-Tuning is a method of fine-tuning a trained model by incorporating both pre-trained and task-specific objectives, balancing the preservation of pre-trained knowledge with adaptation to new tasks. \cite{dong2022leveragingpseudolabeleddataimprove} applied mixed-tuning by training on combined real (primary) and pseudo-labeled (secondary) datasets, ensuring that the larger pseudo-labeled dataset did not dominate the training process. This method improved the model’s performance on diversified data, enhancing generalization.

\subsection{Model Tuning and Distillation}
Cross-Modal Distillation refers to the transfer of knowledge across different modalities, such as text and speech, using a teacher-student framework. This enables models to leverage better modality-specific representations to enhance cross-modal translation tasks \cite{sarkar2023xkdcrossmodalknowledgedistillation}. In \cite{huang2023avtranspeechaudiovisualrobustspeechtospeech}, Cross-Modal Distillation is applied to transfer knowledge from a simple audio-only S2ST model to the AV-TranSpeech model, making it more effective in low-resource settings. The use of additional audio data sources further improves performance, given the limited availability of audio-visual data. Cross-Modal Embedding Alignment, on the other hand, pools features from different modalities, like image and speech, into a shared embedding space. This alignment is essential for tasks such as TTS synthesis and cross-modal retrieval. \cite{duquenne2022tmodulestranslationmoduleszeroshot} uses this technique by aligning speech encoders with LASER text embeddings, fine-tuning the XLSR multilingual speech encoder with MSE loss to map speech embeddings to corresponding text embeddings. This bridges the representation gap between speech and text modalities, enabling zero-shot speech-to-text translation. Teacher-Student Training, a form of model distillation, involves training a smaller or simpler student model to imitate the predictions of a teacher model. \cite{duquenne2022tmodulestranslationmoduleszeroshot} utilizes this method to align language embeddings, where new language-specific text encoders, or "students," are trained from the LASER encoder as a "teacher," producing aligned embeddings within the LASER space. This increases cross-lingual compatibility and improves zero-shot translation effectiveness, especially for low-resource languages like Japanese to English. Prompt Tuning is a technique that adapts pre-trained models to new tasks by modifying the input prompts rather than altering the internal parameters of the model \cite{lester2021powerscaleparameterefficientprompt}. \cite{dong2022leveragingpseudolabeleddataimprove} employs prompt tuning to enhance the model's adaptability across various data sources. By using category-specific prompts, the model distinguishes primary from secondary data at the input end, leading to better inferences over authentic S2ST data through customized embeddings.

\subsection{Translation and Augmentation for Multimodality}
Back-translation is a technique that generates synthetic source-target pairs by translating a target text back into the source language, increasing the available training data for machine translation models and improving translation quality, particularly in low-resource settings. In \cite{nachmani2024translatotron3speechspeech}, back-translation is used during the training phase to improve translation ability. The source language is pseudo-translated into the target language before being decoded back into the source language. This process reduces the differences between embeddings in the translations, thereby enhancing the multilinguality of the latent space and improving translation performance. Pseudo Translation Labeling (PTL) involves creating artificial labels for training by using the outputs of a previously trained model. PTL is particularly effective in enhancing the results of low-resource machine translation tasks \cite{hsu2023pseudolabeltrainingmodelinertia}. \cite{dong2022leveragingpseudolabeleddataimprove} introduces PTL as a solution for sparse data in S2ST, where pseudo-labeled data is created by translating ASR data into pseudo-labeled ST data, which is then passed through a Text-to-Speech system to produce pseudo S2ST data. This increases the size of the training dataset, allowing for pretraining on a larger dataset before fine-tuning on more realistic S2ST data.

\subsection{Sequence Modeling and Generation}
Beam Search is a decoding algorithm that explores a number of candidate sequences at each step and selects the best possible output, commonly used in machine translation and speech synthesis to improve the quality of sequences. In \cite{inaguma2023unitytwopassdirectspeechtospeech}, Beam Search is applied during the inference phase to enhance translation accuracy using a two-pass approach. The first pass expands the beam size to generate multiple possible text translations, while the second pass uses greedy decoding or a reduced beam size to polish the final sequence of discrete elements. \cite{zhang2020uwspeechspeechspeechtranslation} also applies Beam Search in UWSpeech to generate high-probability token sequences from source speech, improving translation quality by exploring multiple hypotheses and selecting the most probable sequence. A Masked Autoencoder works by masking portions of the input data, such as speech or text, and training the model to predict those masked portions. Its self-supervised nature allows the model to learn durable features based on contextual information, which are useful for downstream tasks \cite{germain2015mademaskedautoencoderdistribution}. \cite{nachmani2024translatotron3speechspeech} employs the Masked Autoencoder during the preliminary training phase of Translatotron 3, using reconstruction loss to achieve strong representations. This technique encodes the input into a multilingual embedding framework, while SpecAugment is used as a data augmentation method to help the model generalize and emphasize key parts of the input, preparing it to handle both source and target languages effectively.

\section{Challenges}

\subsection{Data Scarcity and Augmentation}

A central challenge with the architecture of S2ST systems is that there is scarcity of parallel data available for training \cite{tjandra2019speechtospeechtranslationuntranscribedunknown}. To address this, researchers rely on creating artificial data to train S2ST from unlabeled text which somewhat removes shortages in data. Acoustic augmentation techniques enhance the appeal and diversity of synthetic data and model precision and robustness of the model because of cross-lingual embeddings and masked autoencoders (MUSE) \cite{nachmani2024translatotron3speechspeech}.

\subsection{Error Propagation and Model Efficiency}

Error propagation in cascade systems is a major problem where errors occur during the speech recognition stage and propagate to all the later stages of translation. Direct translation models avoid this since it goes without the intermediate text representation; thus accuracy is improved. The use of discrete units has also been an added advantage of serving a compact representation that is less prone to errors \cite{inaguma2023unitytwopassdirectspeechtospeech}. Optimizing the architecture of the speech synthesis has also ensured robustness and efficiency of its accent and dialect handling.

\subsection{Adaptation to Diverse Speech Patterns}

The differences in gender, role, and context affect S2ST systems to adapt themselves using flexibility in changing speech characteristics. Such parameters alter speech rates and pose difficulties in real-time communication, primarily when latency complications emerge in the model architecture \cite{hayakawa-etal-2018-speech}. Researchers emphasize models that can easily adapt these various patterns of speech while reducing latency for effective interaction between the interacting parties \cite{akira17_interspeech} \cite{akira16_interspeech}. Paralinguistic properties is another area of which a translation model needs to stress \cite{akira16_interspeech}.

\subsection{Overcoming Data Scarcity with Multimodal Learning}

The researchers have applied multimodal learning frameworks addressing the availability of datasets concerning the low-resource languages \cite{jia2022cvsscorpusmassivelymultilingual}. The authors have proposed self-supervised pre-training over large datasets as well as synthetic data generation by TTS systems. Such approaches combined with fine-tuning over real-world data triggered tremendous improvement in quality translation regarding low-resource languages. For improving the performance of the system over such contexts, the researchers further claim weakly supervised machine translation data \cite{dong2022leveragingpseudolabeleddataimprove}.

\subsection{Managing Speaker Variability}

The prominent challenge with S2ST models is speaker pronunciation variability and vocal characteristics, which may hamper translation accuracy \cite{zhang2020uwspeechspeechspeechtranslation}. Accordingly, several acoustic enhancements as well as self-supervised learning strategies have been developed to overcome this challenge and bolster generalization. Techniques for voice preservation are required in such a way that the translated information acquired by it captures the identity of the speaker even when speaking in another language \cite{nguyen2022improvingspeechtospeechtranslationunlabeled}.

\subsection{Data Augmentation and Self-Supervised Learning}

Most techniques involve data augmentation techniques in conjunction with self-supervised learning for data augmentation. For example, data synthesis, when two ASR and MT systems are combined, improves the quality of translations in low-resource languages \cite{dong2022leveragingpseudolabeleddataimprove}. Synthetic S2ST data generation and fine-tuning on real-world datasets advance the general performance \cite{duquenne2022tmodulestranslationmoduleszeroshot}.

\subsection{Speech Rate Adaptation and Latency Issues}

Speaker roles, context, and gender variations render the task of speech rate adaptation difficult for S2ST models \cite{akira17_interspeech} \cite{hayakawa-etal-2018-speech} \cite{jia2019directspeechtospeechtranslationsequencetosequence}. Those then would clash with real-time communication especially in latency-prone systems \cite{akira17_interspeech}. The remedy to such difficulties requires models that adapt to changing speech rates without latency bottlenecks \cite{jia2019directspeechtospeechtranslationsequencetosequence}.

\subsection{Handling Phonetic and Syntactic Differences}

The challenge is to handle phonetic and syntactic variations between languages, avoiding intermediate text representations. Researchers propose several robust self-supervised learning methods, synthetic training data, and auxiliary tasks to enhance the quality of translation \cite{DBLP:conf/ococosda/AryaAMP22}. This kind of solution really improved management in speech translation.

\subsection{Model Architectures for Long Sequences}

The greatest significance of S2ST systems lies in modeling long speech sequences with strong generation without either over-generation or under-generation. This problem has overcome with novel model architectures with joint training objectives, a lot of progress being made on the quality of improving translation without losing voice characteristics \cite{nguyen2022improvingspeechtospeechtranslationunlabeled}.

\subsection{Addressing Generalization and Zero-Shot Learning}

Generalization in zero-shot scenarios has always been the biggest challenge-the one, which requires adaptation in unseen styles and speakers using minimal amounts of training data. Several methods for style and semantic content disentanglement have been proposed that permit the model to process both independently, thereby enhancing its generalization capabilities in zero-shot contexts \cite{li2022textlessdirectspeechtospeechtranslation}.

\section{Future Directions}

The reviewed literature suggests a few promising approaches towards the future development of S2ST and multilingual TTS. Some of the most salient focus areas comprise expanding language coverage, improving model architecture, maxing out data usage, and optimizing systems for practical use in the real world. Delineated in the following sections are prospective future directions taking findings from such studies.

\subsection{Expansion of Language Coverage and Multilingual Training}

Much attention in many studies have been put on the necessity to make S2ST systems much more affordable for a higher number of languages, some of which are low-resource, and even unwritten. Improvement can be done through transfer learning, multilingual training approaches, and the development of more unified encoders and decoders that can efficiently deal with multiple language pairs and dialects \cite{zhu2023diffs2utsemanticpreservingdiffusion}, \cite{wei2022jointpretrainingspeechbilingual}, \cite{inaguma2023unitytwopassdirectspeechtospeech}, \cite{Mingote_2023}, \cite{akira16_interspeech}, \cite{jia2022cvsscorpusmassivelymultilingual}, \cite{dong2022leveragingpseudolabeleddataimprove}, \cite{DBLP:conf/ococosda/AryaAMP22}, \cite{nguyen2022improvingspeechtospeechtranslationunlabeled}. Addressing language-specific challenges in the form of phonetic, syntactic, and semantic nuances will further strengthen model robustness and adaptability across different linguistic contexts \cite{zhang2020uwspeechspeechspeechtranslation}, \cite{duquenne2022tmodulestranslationmoduleszeroshot}.

\subsection{Enhanced Data and Corpus Development}

The future work may have large and diverse datasets, high-quality human-annotated data, and synthetic data for low-resource languages. It would further help in refining the S2ST and TTS models so that they can generalize across a wider range of linguistic and acoustic variations, as demonstrated in Reference \cite{huang2023avtranspeechaudiovisualrobustspeechtospeech}, \cite{Mingote_2023}, \cite{akira16_interspeech}, \cite{jia2022leveragingunsupervisedweaklysuperviseddata}, \cite{lee2022directspeechtospeechtranslationdiscrete}. Data augmentation techniques, including multimodal integration, could also help in increasing the depth and diversity of training data. This would provide more contextual information for accurate translation \cite{nachmani2024translatotron3speechspeech}.

\subsection{Optimizing Model Architecture and Efficiency}

This has been especially crucial for real-time applications that involve better efficiency and scalability of the models. Envisaged future improvements would involve further research in new architectures that could possibly reduce the computations and improve processing speeds. These would include novel, advanced TTS models and vocoders optimized for fast speech-to-speech translation capabilities \cite{shi2023enhancingspeechtospeechtranslationmultiple}, \cite{yan2023espnetstv2multipurposespokenlanguage}, \cite{jia2019directspeechtospeechtranslationsequencetosequence}, \cite{DBLP:conf/ococosda/AryaAMP22}. Moreover, with fewer generation steps and further optimizations for direct translation, these models will find increased usage in mobile and embedded systems, among others \cite{jia2022translatotron2highqualitydirect}.

\subsection{Pre-Training and Transfer Learning Techniques}

New advances in pre-training provide great hope for improving the performance of models, especially in domains where available labeled data is limited. Future work will focus on pre-training methods that better capture complex frameworks of language and leverage cross-linguistic similarity to improve model generalization  \cite{wu2016googlesneuralmachinetranslation}, \cite{jia2022leveragingunsupervisedweaklysuperviseddata}, \cite{popuri2022enhanceddirectspeechtospeechtranslation}. Methods of multilingual and multitask pre-training may enable the consistent performance of models across languages but also enable fine-tuning to specific tasks within S2ST and TTS systems \cite{nguyen2022improvingspeechtospeechtranslationunlabeled}.

\subsection{Real-Time Applications and Latency Reduction}

Future research may be directed at low latency reductions without any loss of accuracy. These comprise low latency architecture models, efficient algorithms for the training process, and also optimizations in the hardware that improve the processing speed in the interactive application scenarios  \cite{fang2023daspeechdirectedacyclictransformer}, \cite{nguyen2022improvingspeechtospeechtranslationunlabeled}, \cite{lee2022directspeechtospeechtranslationdiscrete}. Low latency performance is vital to make possible the wider use of S2ST systems within the international communication, business, and health fields \cite{hayakawa-etal-2018-speech}, \cite{jia2022translatotron2highqualitydirect}.

\subsection{Improving Model Robustness and Noise Resilience}

One of the main requirements is that all models' formulations should be robust in their performance under various changes of noise levels and differences in the quality of input. This challenge can be overcome by using noise-resilient training methodologies and adaptive feedback systems guaranteeing model efficiency under different conditions, like noisy environments and different speaker characteristics  \cite{huang2023avtranspeechaudiovisualrobustspeechtospeech}, \cite{inaguma2023unitytwopassdirectspeechtospeech}, \cite{tjandra2019speechtospeechtranslationuntranscribedunknown}, \cite{nachmani2024translatotron3speechspeech}. Robustness enhancement will determine increasing accessibility and user satisfaction in realistic scenarios \cite{fang2023daspeechdirectedacyclictransformer}, \cite{zhang2020uwspeechspeechspeechtranslation}.

\subsection{Integration of Paralinguistic and Non-Linguistic Features}

Various studies indicate that next-generation systems should retain more paralinguistic and extralinguistic features like intonation, emotional expression, and speaker identity to gain more nativelike translations. Techniques aimed toward the production of such features are worthy of investigation in order to render the output more expressive and human-like-like interacting  \cite{tjandra2019speechtospeechtranslationuntranscribedunknown}, \cite{jia2019directspeechtospeechtranslationsequencetosequence}, \cite{jia2022leveragingunsupervisedweaklysuperviseddata}, \cite{popuri2022enhanceddirectspeechtospeechtranslation}. Finally, the further developed style transfer techniques would enhance the emotional and stylistic qualities of synthesized speech to become more arousing and articulate for the listener \cite{song2023styles2stzeroshotstyletransfer}.

\subsection{Evaluation Metrics and Benchmark Development}
Advanced metrics and benchmarks have to be used for effective benchmarking. Studies have demonstrated that the use of human evaluation along with automated metrics results in a better estimation of the quality and naturalness of the translation \cite{huang2023avtranspeechaudiovisualrobustspeechtospeech}, \cite{inaguma2023unitytwopassdirectspeechtospeech}, \cite{Quintana2018ADS}. Additional variables that relate to a specific context, like speech rate or noise, can further enhance the diagnostics of the landmarks  \cite{akira17_interspeech}, \cite{jia2019directspeechtospeechtranslationsequencetosequence}.
\\\\
Thus, the outlined future directions indicate that the discipline had matured so that the generalized expansion of the coverage of languages, model robustness improvement, and optimization of efficiency would enable more inclusive and practical applications to emerge. Focusing on these aspects, the forthcoming generation of S2ST and TTS systems will better fulfill the diversified needs of users in a more and more globalized and multilingual environment.

\section{Conclusion}
Speech-to-Speech Translation (S2ST) has come a long way, first moving away from cascade models and now more towards direct end-to-end systems. On the positive side, cascade models allow modularity; however, they suffer from error propagation and increased latency. Direct S2ST models, such as Translatotron 2 and UnitY, simplify the translation process by circumventing intermediate text, lowering latency, and maintaining speaker identity. These models, however, suffer from a dearth of data and the demand for large-scale computational resources—especially for low-resource and linguistically diverse languages.
\\\\
The difficulties of these tasks notwithstanding, advances in self-supervised learning, data augmentation, and multimodal approaches have greatly improved the quality and robustness of translations. Models integrating audio-visual features enhance real-time translation and expand multilingual capabilities. Future research is urged to give much attention to the development of comprehensive datasets, efficient architectures, and ethical frameworks to ensure that S2ST technologies are applied fairly in rich linguistic and cultural landscapes.

\bibliographystyle{unsrt}
\bibliography{ReviewPaper}

\begin{thebibliography}{100}

\bibitem{communication2023seamlessmultilingualexpressivestreaming}
Seamless Communication, Loïc Barrault, Yu-An Chung, Mariano~Coria Meglioli, David Dale, Ning Dong, Mark Duppenthaler, Paul-Ambroise Duquenne, Brian Ellis, Hady Elsahar, Justin Haaheim, John Hoffman, Min-Jae Hwang, Hirofumi Inaguma, Christopher Klaiber, Ilia Kulikov, Pengwei Li, Daniel Licht, Jean Maillard, Ruslan Mavlyutov, Alice Rakotoarison, Kaushik~Ram Sadagopan, Abinesh Ramakrishnan, Tuan Tran, Guillaume Wenzek, Yilin Yang, Ethan Ye, Ivan Evtimov, Pierre Fernandez, Cynthia Gao, Prangthip Hansanti, Elahe Kalbassi, Amanda Kallet, Artyom Kozhevnikov, Gabriel~Mejia Gonzalez, Robin~San Roman, Christophe Touret, Corinne Wong, Carleigh Wood, Bokai Yu, Pierre Andrews, Can Balioglu, Peng-Jen Chen, Marta~R. Costa-jussà, Maha Elbayad, Hongyu Gong, Francisco Guzmán, Kevin Heffernan, Somya Jain, Justine Kao, Ann Lee, Xutai Ma, Alex Mourachko, Benjamin Peloquin, Juan Pino, Sravya Popuri, Christophe Ropers, Safiyyah Saleem, Holger Schwenk, Anna Sun, Paden Tomasello, Changhan Wang, Jeff Wang, Skyler Wang, and Mary
  Williamson.
\newblock Seamless: Multilingual expressive and streaming speech translation, 2023.

\bibitem{jia2022translatotron2highqualitydirect}
Ye~Jia, Michelle~Tadmor Ramanovich, Tal Remez, and Roi Pomerantz.
\newblock Translatotron 2: High-quality direct speech-to-speech translation with voice preservation, 2022.

\bibitem{le2024transvipspeechspeechtranslation}
Chenyang Le, Yao Qian, Dongmei Wang, Long Zhou, Shujie Liu, Xiaofei Wang, Midia Yousefi, Yanmin Qian, Jinyu Li, Sheng Zhao, and Michael Zeng.
\newblock Transvip: Speech to speech translation system with voice and isochrony preservation, 2024.

\bibitem{ijcat}
Sandeep Dhawan.
\newblock Speech to speech translation: Challenges and future, 2022.
\newblock [Accessed 19-11-2024].

\bibitem{wu2016googlesneuralmachinetranslation}
Yonghui Wu, Mike Schuster, Zhifeng Chen, Quoc~V. Le, Mohammad Norouzi, Wolfgang Macherey, Maxim Krikun, Yuan Cao, Qin Gao, Klaus Macherey, Jeff Klingner, Apurva Shah, Melvin Johnson, Xiaobing Liu, Łukasz Kaiser, Stephan Gouws, Yoshikiyo Kato, Taku Kudo, Hideto Kazawa, Keith Stevens, George Kurian, Nishant Patil, Wei Wang, Cliff Young, Jason Smith, Jason Riesa, Alex Rudnick, Oriol Vinyals, Greg Corrado, Macduff Hughes, and Jeffrey Dean.
\newblock Google's neural machine translation system: Bridging the gap between human and machine translation, 2016.

\bibitem{li-etal-2017-translating}
Hongzheng Li, Philippe Langlais, and Yaohong Jin.
\newblock Translating implicit discourse connectives based on cross-lingual annotation and alignment.
\newblock In Bonnie Webber, Andrei Popescu-Belis, and J{\"o}rg Tiedemann, editors, {\em Proceedings of the Third Workshop on Discourse in Machine Translation}, pages 93--98, Copenhagen, Denmark, September 2017. Association for Computational Linguistics.

\bibitem{app11104380}
You-Sik Hong, Chang-Pyoung Han, and Seong-Soo Cho.
\newblock Level-based learning algorithm based on the difficulty level of the test problem.
\newblock {\em Applied Sciences}, 11(10), 2021.

\bibitem{dasgupta2020improvinglocalidentifiabilityprobabilistic}
Shib~Sankar Dasgupta, Michael Boratko, Dongxu Zhang, Luke Vilnis, Xiang~Lorraine Li, and Andrew McCallum.
\newblock Improving local identifiability in probabilistic box embeddings, 2020.

\bibitem{jia2019directspeechtospeechtranslationsequencetosequence}
Ye~Jia, Ron~J. Weiss, Fadi Biadsy, Wolfgang Macherey, Melvin Johnson, Zhifeng Chen, and Yonghui Wu.
\newblock Direct speech-to-speech translation with a sequence-to-sequence model, 2019.

\bibitem{pratapa-etal-2018-language}
Adithya Pratapa, Gayatri Bhat, Monojit Choudhury, Sunayana Sitaram, Sandipan Dandapat, and Kalika Bali.
\newblock Language modeling for code-mixing: The role of linguistic theory based synthetic data.
\newblock In Iryna Gurevych and Yusuke Miyao, editors, {\em Proceedings of the 56th Annual Meeting of the Association for Computational Linguistics (Volume 1: Long Papers)}, pages 1543--1553, Melbourne, Australia, July 2018. Association for Computational Linguistics.

\bibitem{communication2023seamlessm4tmassivelymultilingual}
Seamless Communication, Loïc Barrault, Yu-An Chung, Mariano~Cora Meglioli, David Dale, Ning Dong, Paul-Ambroise Duquenne, Hady Elsahar, Hongyu Gong, Kevin Heffernan, John Hoffman, Christopher Klaiber, Pengwei Li, Daniel Licht, Jean Maillard, Alice Rakotoarison, Kaushik~Ram Sadagopan, Guillaume Wenzek, Ethan Ye, Bapi Akula, Peng-Jen Chen, Naji~El Hachem, Brian Ellis, Gabriel~Mejia Gonzalez, Justin Haaheim, Prangthip Hansanti, Russ Howes, Bernie Huang, Min-Jae Hwang, Hirofumi Inaguma, Somya Jain, Elahe Kalbassi, Amanda Kallet, Ilia Kulikov, Janice Lam, Daniel Li, Xutai Ma, Ruslan Mavlyutov, Benjamin Peloquin, Mohamed Ramadan, Abinesh Ramakrishnan, Anna Sun, Kevin Tran, Tuan Tran, Igor Tufanov, Vish Vogeti, Carleigh Wood, Yilin Yang, Bokai Yu, Pierre Andrews, Can Balioglu, Marta~R. Costa-jussà, Onur Celebi, Maha Elbayad, Cynthia Gao, Francisco Guzmán, Justine Kao, Ann Lee, Alexandre Mourachko, Juan Pino, Sravya Popuri, Christophe Ropers, Safiyyah Saleem, Holger Schwenk, Paden Tomasello, Changhan Wang, Jeff
  Wang, and Skyler Wang.
\newblock Seamlessm4t: Massively multilingual \& multimodal machine translation, 2023.

\bibitem{10.1007/978-3-031-48312-7_21}
Lalaram Arya, Amartya Roy~Chowdhury, and S.~R.~Mahadeva Prasanna.
\newblock Direct vs cascaded speech-to-speech translation using transformer.
\newblock In Alexey Karpov, K.~Samudravijaya, K.~T. Deepak, Rajesh~M. Hegde, Shyam~S. Agrawal, and S.~R.~Mahadeva Prasanna, editors, {\em Speech and Computer}, pages 258--270, Cham, 2023. Springer Nature Switzerland.

\bibitem{han2024speechqeestimatingqualitydirect}
HyoJung Han, Kevin Duh, and Marine Carpuat.
\newblock Speechqe: Estimating the quality of direct speech translation, 2024.

\bibitem{xu2023recentadvancesdirectspeechtotext}
Chen Xu, Rong Ye, Qianqian Dong, Chengqi Zhao, Tom Ko, Mingxuan Wang, Tong Xiao, and Jingbo Zhu.
\newblock Recent advances in direct speech-to-text translation, 2023.

\bibitem{key}
Sireesh~Haang Limbu.
\newblock {D}irect {S}peech to {S}peech {T}ranslation {U}sing {M}achine {L}earning, 2020.
\newblock [Accessed 20-11-2024].

\bibitem{lee2022directspeechtospeechtranslationdiscrete}
Ann Lee, Peng-Jen Chen, Changhan Wang, Jiatao Gu, Sravya Popuri, Xutai Ma, Adam Polyak, Yossi Adi, Qing He, Yun Tang, Juan Pino, and Wei-Ning Hsu.
\newblock Direct speech-to-speech translation with discrete units, 2022.

\bibitem{inaguma2023unitytwopassdirectspeechtospeech}
Hirofumi Inaguma, Sravya Popuri, Ilia Kulikov, Peng-Jen Chen, Changhan Wang, Yu-An Chung, Yun Tang, Ann Lee, Shinji Watanabe, and Juan Pino.
\newblock Unity: Two-pass direct speech-to-speech translation with discrete units, 2023.

\bibitem{GoogleAIBlog}
Ron~Weiss Ye~Jia.
\newblock Introducing translatotron: An end-to-end speech-to-speech translation model.

\bibitem{wei2022jointpretrainingspeechbilingual}
Kun Wei, Long Zhou, Ziqiang Zhang, Liping Chen, Shujie Liu, Lei He, Jinyu Li, and Furu Wei.
\newblock Joint pre-training with speech and bilingual text for direct speech to speech translation, 2022.

\bibitem{Al-Tarawneh2024}
Alalddin Al-Tarawneh.
\newblock Bridging languages and numbers: Exploring the intersection of translation studies and mathematics.
\newblock {\em Applied Mathematics \& Information Sciences}, 18(3):513--519, 2024.

\bibitem{PullumKornai2003}
Geoffrey~K. Pullum and András Kornai.
\newblock Mathematical linguistics.
\newblock In William Frawley, editor, {\em The Oxford International Encyclopedia of Linguistics}, pages 17--20. Oxford University Press, 2nd edition, 2003.

\bibitem{Rahmawati2020}
Dwi Rahmawati and Rahmad~B. Anwar.
\newblock Translation of mathematical representation: Characteristics of verbal representation unpacking.
\newblock {\em Journal of Education and Learning (EduLearn)}, 14(2):163--170, 2020.

\bibitem{10.3115/1073083.1073135}
Kishore Papineni, Salim Roukos, Todd Ward, and Wei-Jing Zhu.
\newblock Bleu: a method for automatic evaluation of machine translation.
\newblock In {\em Proceedings of the 40th Annual Meeting on Association for Computational Linguistics}, ACL '02, page 311–318, USA, 2002. Association for Computational Linguistics.

\bibitem{DBLP:journals/corr/abs-1804-08771}
Matt Post.
\newblock A call for clarity in reporting {BLEU} scores.
\newblock {\em CoRR}, abs/1804.08771, 2018.

\bibitem{Kornai2007}
András Kornai.
\newblock {\em Mathematical Linguistics}.
\newblock Springer, 2007.

\bibitem{ijcaonline}
Sumanlata~Gautam Mahak~Dureja.
\newblock Speech-to-speech translation: A review, 2015.
\newblock [Accessed 29-11-2024].

\bibitem{arxiv1}
Chandresh Kumar~Maurya Mahendra~Gupta, Maitreyee~Dutta.
\newblock Direct speech-to-speech neural machine translation: A survey, 2024.
\newblock [Accessed 29-11-2024].

\bibitem{kaur2024directpunjabienglishspeech}
Prabhjot Kaur, L.~Andrew~M. Bush, and Weisong Shi.
\newblock Direct punjabi to english speech translation using discrete units, 2024.

\bibitem{fang2024achievehighqualitydirectspeechtospeech}
Qingkai Fang, Shaolei Zhang, Zhengrui Ma, Min Zhang, and Yang Feng.
\newblock Can we achieve high-quality direct speech-to-speech translation without parallel speech data?, 2024.

\bibitem{carolefrench}
TALA METALOM~Diane Carole and YENKE~Blaise Omer.
\newblock French-fulfulde textless and cascading speech translation: towards a dual architecture.
\newblock 2024.

\bibitem{liu2023multi}
Rouhe Liu, Yue Zhao, and Xiaona Xu.
\newblock Multi-task self-supervised learning based tibetan-chinese speech-to-speech translation.
\newblock In {\em 2023 International Conference on Asian Language Processing (IALP)}, pages 45--49. IEEE, 2023.

\bibitem{nachmani2024translatotron}
Eliya Nachmani, Alon Levkovitch, Yifan Ding, Chulayuth Asawaroengchai, Heiga Zen, and Michelle~Tadmor Ramanovich.
\newblock Translatotron 3: Speech to speech translation with monolingual data.
\newblock In {\em ICASSP 2024-2024 IEEE International Conference on Acoustics, Speech and Signal Processing (ICASSP)}, pages 10686--10690. IEEE, 2024.

\bibitem{lee2022direct}
Ann Lee, Peng-Jen Chen, Changhan Wang, Jiatao Gu, Sravya Popuri, Xutai Ma, Adam Polyak, Yossi Adi, Qing He, Yun Tang, et~al.
\newblock Direct speech-to-speech translation with discrete units.
\newblock {\em arXiv preprint arXiv:2107.05604}, 2021.

\bibitem{zhang2021uwspeech}
Chen Zhang, Xu~Tan, Yi~Ren, Tao Qin, Kejun Zhang, and Tie-Yan Liu.
\newblock Uwspeech: Speech to speech translation for unwritten languages.
\newblock In {\em Proceedings of the AAAI Conference on Artificial Intelligence}, volume~35, pages 14319--14327, 2021.

\bibitem{jia2019direct}
Ye~Jia, Ron~J Weiss, Fadi Biadsy, Wolfgang Macherey, Melvin Johnson, Zhifeng Chen, and Yonghui Wu.
\newblock Direct speech-to-speech translation with a sequence-to-sequence model.
\newblock {\em arXiv preprint arXiv:1904.06037}, 2019.

\bibitem{10229412}
Hsiao-Chuan Liu, Min-Yuh Day, and Chih-Chien Wang.
\newblock Speech-to-speech low-resource translation.
\newblock In {\em 2023 IEEE 24th International Conference on Information Reuse and Integration for Data Science (IRI)}, pages 91--95, 2023.

\bibitem{igntu}
Ayushi Trivedi, Navya Pant, Pinal Shah, Simran Sonik, and Supriya Agrawal.
\newblock Speech to text and text to speech recognition systems-areview, 2018.
\newblock [Accessed 29-11-2024].

\bibitem{book}
Jaffer Khan, Alamelu Raman, Nithya Sambamoorthy, and Kanniga Prashanth.
\newblock {\em Research Methodology (Methods, Approaches And Techniques)}.
\newblock 09 2023.

\bibitem{articleMOS}
Robert Streijl and David Hands.
\newblock Mean opinion score (mos) revisited: methods and applications, limitations and alternatives.
\newblock {\em Multimedia Systems}, 22:213--227, 03 2016.

\bibitem{wang2024uncertaintyawaremeanopinionscore}
Hui Wang, Shiwan Zhao, Jiaming Zhou, Xiguang Zheng, Haoqin Sun, Xuechen Wang, and Yong Qin.
\newblock Uncertainty-aware mean opinion score prediction, 2024.

\bibitem{latency}
Ali~Safari Khatouni, Francesca Soro, and Danilo Giordano.
\newblock A machine learning application for latency prediction in operational 4g networks.
\newblock In {\em 2019 IFIP/IEEE Symposium on Integrated Network and Service Management (IM)}, pages 71--74, 2019.

\bibitem{shi2023enhancingspeechtospeechtranslationmultiple}
Jiatong Shi, Yun Tang, Ann Lee, Hirofumi Inaguma, Changhan Wang, Juan Pino, and Shinji Watanabe.
\newblock Enhancing speech-to-speech translation with multiple tts targets, 2023.

\bibitem{zhang2020uwspeechspeechspeechtranslation}
Chen Zhang, Xu~Tan, Yi~Ren, Tao Qin, Kejun Zhang, and Tie-Yan Liu.
\newblock Uwspeech: Speech to speech translation for unwritten languages, 2020.

\bibitem{li2022textlessdirectspeechtospeechtranslation}
Xinjian Li, Ye~Jia, and Chung-Cheng Chiu.
\newblock Textless direct speech-to-speech translation with discrete speech representation, 2022.

\bibitem{zhu2023diffs2utsemanticpreservingdiffusion}
Yongxin Zhu, Zhujin Gao, Xinyuan Zhou, Zhongyi Ye, and Linli Xu.
\newblock Diffs2ut: A semantic preserving diffusion model for textless direct speech-to-speech translation, 2023.

\bibitem{nguyen2022improvingspeechtospeechtranslationunlabeled}
Xuan-Phi Nguyen, Sravya Popuri, Changhan Wang, Yun Tang, Ilia Kulikov, and Hongyu Gong.
\newblock Improving speech-to-speech translation through unlabeled text, 2022.

\bibitem{popuri2022enhanceddirectspeechtospeechtranslation}
Sravya Popuri, Peng-Jen Chen, Changhan Wang, Juan Pino, Yossi Adi, Jiatao Gu, Wei-Ning Hsu, and Ann Lee.
\newblock Enhanced direct speech-to-speech translation using self-supervised pre-training and data augmentation, 2022.

\bibitem{fang2023daspeechdirectedacyclictransformer}
Qingkai Fang, Yan Zhou, and Yang Feng.
\newblock Daspeech: Directed acyclic transformer for fast and high-quality speech-to-speech translation, 2023.

\bibitem{yan2023espnetstv2multipurposespokenlanguage}
Brian Yan, Jiatong Shi, Yun Tang, Hirofumi Inaguma, Yifan Peng, Siddharth Dalmia, Peter Polák, Patrick Fernandes, Dan Berrebbi, Tomoki Hayashi, Xiaohui Zhang, Zhaoheng Ni, Moto Hira, Soumi Maiti, Juan Pino, and Shinji Watanabe.
\newblock Espnet-st-v2: Multipurpose spoken language translation toolkit, 2023.

\bibitem{nachmani2024translatotron3speechspeech}
Eliya Nachmani, Alon Levkovitch, Yifan Ding, Chulayuth Asawaroengchai, Heiga Zen, and Michelle~Tadmor Ramanovich.
\newblock Translatotron 3: Speech to speech translation with monolingual data, 2024.

\bibitem{duquenne2022tmodulestranslationmoduleszeroshot}
Paul-Ambroise Duquenne, Hongyu Gong, Benoît Sagot, and Holger Schwenk.
\newblock T-modules: Translation modules for zero-shot cross-modal machine translation, 2022.

\bibitem{Mingote_2023}
Victoria Mingote, Pablo Gimeno, Luis Vicente, Sameer Khurana, Antoine Laurent, and Jarod Duret.
\newblock Direct text to speech translation system using acoustic units.
\newblock {\em IEEE Signal Processing Letters}, 30:1262–1266, 2023.

\bibitem{jia2022cvsscorpusmassivelymultilingual}
Ye~Jia, Michelle~Tadmor Ramanovich, Quan Wang, and Heiga Zen.
\newblock Cvss corpus and massively multilingual speech-to-speech translation, 2022.

\bibitem{jia2022leveragingunsupervisedweaklysuperviseddata}
Ye~Jia, Yifan Ding, Ankur Bapna, Colin Cherry, Yu~Zhang, Alexis Conneau, and Nobuyuki Morioka.
\newblock Leveraging unsupervised and weakly-supervised data to improve direct speech-to-speech translation, 2022.

\bibitem{Quintana2018ADS}
Manuel Quintana and Miguel Bernal.
\newblock A direct speech-to-speech neural network methodology for spanish-english translation.
\newblock {\em EAI Endorsed Trans. Energy Web}, 7:e4, 2018.

\bibitem{song2023styles2stzeroshotstyletransfer}
Kun Song, Yi~Ren, Yi~Lei, Chunfeng Wang, Kun Wei, Lei Xie, Xiang Yin, and Zejun Ma.
\newblock Styles2st: Zero-shot style transfer for direct speech-to-speech translation, 2023.

\bibitem{wang-etal-2021-voxpopuli}
Changhan Wang, Morgane Riviere, Ann Lee, Anne Wu, Chaitanya Talnikar, Daniel Haziza, Mary Williamson, Juan Pino, and Emmanuel Dupoux.
\newblock {V}ox{P}opuli: A large-scale multilingual speech corpus for representation learning, semi-supervised learning and interpretation.
\newblock In Chengqing Zong, Fei Xia, Wenjie Li, and Roberto Navigli, editors, {\em Proceedings of the 59th Annual Meeting of the Association for Computational Linguistics and the 11th International Joint Conference on Natural Language Processing (Volume 1: Long Papers)}, pages 993--1003, Online, August 2021. Association for Computational Linguistics.

\bibitem{post-etal-2013-improved}
Matt Post, Gaurav Kumar, Adam Lopez, Damianos Karakos, Chris Callison-Burch, and Sanjeev Khudanpur.
\newblock Improved speech-to-text translation with the fisher and callhome {S}panish-{E}nglish speech translation corpus.
\newblock In Joy~Ying Zhang, editor, {\em Proceedings of the 10th International Workshop on Spoken Language Translation: Papers}, Heidelberg, Germany, December 5-6 2013.

\bibitem{dong2022leveragingpseudolabeleddataimprove}
Qianqian Dong, Fengpeng Yue, Tom Ko, Mingxuan Wang, Qibing Bai, and Yu~Zhang.
\newblock Leveraging pseudo-labeled data to improve direct speech-to-speech translation, 2022.

\bibitem{iranzosánchez2020europarlstmultilingualcorpusspeech}
Javier Iranzo-Sánchez, Joan~Albert Silvestre-Cerdà, Javier Jorge, Nahuel Roselló, Adrià Giménez, Albert Sanchis, Jorge Civera, and Alfons Juan.
\newblock Europarl-st: A multilingual corpus for speech translation of parliamentary debates, 2020.

\bibitem{jia2022introducing}
Ye~Jia and Michelle~Tadmor Ramanovich.
\newblock Introducing cvss: A massively multilingual speech-to-speech translation corpus, April 2022.

\bibitem{jia-etal-2022-cvss}
Ye~Jia, Michelle Tadmor~Ramanovich, Quan Wang, and Heiga Zen.
\newblock {CVSS} corpus and massively multilingual speech-to-speech translation.
\newblock In Nicoletta Calzolari, Fr{\'e}d{\'e}ric B{\'e}chet, Philippe Blache, Khalid Choukri, Christopher Cieri, Thierry Declerck, Sara Goggi, Hitoshi Isahara, Bente Maegaard, Joseph Mariani, H{\'e}l{\`e}ne Mazo, Jan Odijk, and Stelios Piperidis, editors, {\em Proceedings of the Thirteenth Language Resources and Evaluation Conference}, pages 6691--6703, Marseille, France, June 2022. European Language Resources Association.

\bibitem{huang2023avtranspeechaudiovisualrobustspeechtospeech}
Rongjie Huang, Huadai Liu, Xize Cheng, Yi~Ren, Linjun Li, Zhenhui Ye, Jinzheng He, Lichao Zhang, Jinglin Liu, Xiang Yin, and Zhou Zhao.
\newblock Av-transpeech: Audio-visual robust speech-to-speech translation, 2023.

\bibitem{afouras2018lrs3tedlargescaledatasetvisual}
Triantafyllos Afouras, Joon~Son Chung, and Andrew Zisserman.
\newblock Lrs3-ted: a large-scale dataset for visual speech recognition, 2018.

\bibitem{chen2021gigaspeechevolvingmultidomainasr}
Guoguo Chen, Shuzhou Chai, Guanbo Wang, Jiayu Du, Wei-Qiang Zhang, Chao Weng, Dan Su, Daniel Povey, Jan Trmal, Junbo Zhang, Mingjie Jin, Sanjeev Khudanpur, Shinji Watanabe, Shuaijiang Zhao, Wei Zou, Xiangang Li, Xuchen Yao, Yongqing Wang, Yujun Wang, Zhao You, and Zhiyong Yan.
\newblock Gigaspeech: An evolving, multi-domain asr corpus with 10,000 hours of transcribed audio, 2021.

\bibitem{di-gangi-etal-2019-must}
Mattia~A. Di~Gangi, Roldano Cattoni, Luisa Bentivogli, Matteo Negri, and Marco Turchi.
\newblock {M}u{ST}-{C}: a {M}ultilingual {S}peech {T}ranslation {C}orpus.
\newblock In Jill Burstein, Christy Doran, and Thamar Solorio, editors, {\em Proceedings of the 2019 Conference of the North {A}merican Chapter of the Association for Computational Linguistics: Human Language Technologies, Volume 1 (Long and Short Papers)}, pages 2012--2017, Minneapolis, Minnesota, June 2019. Association for Computational Linguistics.

\bibitem{BTEC}
Genichiro Kikui, Eiichiro Sumita, Toshiyuki Takezawa, and Seiichi Yamamoto.
\newblock Creating corpora for speech-to-speech translation.
\newblock 09 2003.

\bibitem{tjandra2019speechtospeechtranslationuntranscribedunknown}
Andros Tjandra, Sakriani Sakti, and Satoshi Nakamura.
\newblock Speech-to-speech translation between untranscribed unknown languages, 2019.

\bibitem{anderson1991hcrc}
Anne~H. Anderson, Markus Bader, Ellen~G. Bard, Elizabeth Boyle, Gwyneth Doherty, Simon Garrod, Stephen Isard, Jacqueline Kowtko, James McAllister, Jim Miller, Catherine Sotillo, Henry~S. Thompson, and Regina Weinert.
\newblock The hcrc map task corpus.
\newblock {\em Language and Speech}, 34(4):351--366, 1991.

\bibitem{akira16_interspeech}
Hayakawa Akira, Saturnino Luz, and Nick Campbell.
\newblock Talking to a system and talking to a human: A study from a speech-to-speech, machine translation mediated map task.
\newblock In {\em Interspeech 2016}, pages 1422--1426, 2016.

\bibitem{akira17_interspeech}
Hayakawa Akira, Carl Vogel, Saturnino Luz, and Nick Campbell.
\newblock Speech rate comparison when talking to a system and talking to a human: A study from a speech-to-speech, machine translation mediated map task.
\newblock In {\em Interspeech 2017}, pages 3286--3290, 2017.

\bibitem{hayakawa-etal-2018-speech}
Akira Hayakawa, Carl Vogel, Saturnino Luz, and Nick Campbell.
\newblock Speech rate calculations with short utterances: A study from a speech-to-speech, machine translation mediated map task.
\newblock In Nicoletta Calzolari, Khalid Choukri, Christopher Cieri, Thierry Declerck, Sara Goggi, Koiti Hasida, Hitoshi Isahara, Bente Maegaard, Joseph Mariani, H{\'e}l{\`e}ne Mazo, Asuncion Moreno, Jan Odijk, Stelios Piperidis, and Takenobu Tokunaga, editors, {\em Proceedings of the Eleventh International Conference on Language Resources and Evaluation ({LREC} 2018)}, Miyazaki, Japan, May 2018. European Language Resources Association (ELRA).

\bibitem{hayakawa-etal-2016-ilmt}
Akira Hayakawa, Saturnino Luz, Loredana Cerrato, and Nick Campbell.
\newblock The {ILMT}-s2s corpus {\textemdash} a multimodal interlingual map task corpus.
\newblock In Nicoletta Calzolari, Khalid Choukri, Thierry Declerck, Sara Goggi, Marko Grobelnik, Bente Maegaard, Joseph Mariani, Helene Mazo, Asuncion Moreno, Jan Odijk, and Stelios Piperidis, editors, {\em Proceedings of the Tenth International Conference on Language Resources and Evaluation ({LREC}`16)}, pages 605--612, Portoro{\v{z}}, Slovenia, May 2016. European Language Resources Association (ELRA).

\bibitem{wang2020covost2massivelymultilingual}
Changhan Wang, Anne Wu, and Juan Pino.
\newblock Covost 2 and massively multilingual speech-to-text translation, 2020.

\bibitem{salesky2021mtedx}
Elizabeth Salesky, Matthew Wiesner, Jacob Bremerman, Roldano Cattoni, Matteo Negri, Marco Turchi, Douglas~W. Oard, and Matt Post.
\newblock Multilingual tedx corpus for speech recognition and translation.
\newblock In {\em Proceedings of Interspeech}, 2021.

\bibitem{Pratap_2020}
Vineel Pratap, Qiantong Xu, Anuroop Sriram, Gabriel Synnaeve, and Ronan Collobert.
\newblock Mls: A large-scale multilingual dataset for speech research.
\newblock In {\em Interspeech 2020}, interspeech\_2020. ISCA, October 2020.

\bibitem{commonvoice2023}
Mozilla.
\newblock Common voice dataset.
\newblock \url{https://commonvoice.mozilla.org/en/datasets}, 2023.
\newblock Accessed: 2023-10-25.

\bibitem{7178964}
Vassil Panayotov, Guoguo Chen, Daniel Povey, and Sanjeev Khudanpur.
\newblock Librispeech: An asr corpus based on public domain audio books.
\newblock In {\em 2015 IEEE International Conference on Acoustics, Speech and Signal Processing (ICASSP)}, pages 5206--5210, 2015.

\bibitem{Hernandez_2018}
François Hernandez, Vincent Nguyen, Sahar Ghannay, Natalia Tomashenko, and Yannick Estève.
\newblock {\em TED-LIUM 3: Twice as Much Data and Corpus Repartition for Experiments on Speaker Adaptation}, page 198–208.
\newblock Springer International Publishing, 2018.

\bibitem{liu19d_interspeech}
Yuchen Liu, Hao Xiong, Jiajun Zhang, Zhongjun He, Hua Wu, Haifeng Wang, and Chengqing Zong.
\newblock End-to-end speech translation with knowledge distillation.
\newblock In {\em Interspeech 2019}, pages 1128--1132, 2019.

\bibitem{wenzek2019ccnetextractinghighquality}
Guillaume Wenzek, Marie-Anne Lachaux, Alexis Conneau, Vishrav Chaudhary, Francisco Guzmán, Armand Joulin, and Edouard Grave.
\newblock Ccnet: Extracting high quality monolingual datasets from web crawl data, 2019.

\bibitem{DBLP:conf/ococosda/AryaAMP22}
Lalaram Arya, Ayush Agarwal, Jagabandhu Mishra, and S.~R.~Mahadeva Prasanna.
\newblock Analysis of layer-wise training in direct speech to speech translation using {BI-LSTM}.
\newblock In {\em 25th Conference of the Oriental {COCOSDA} International Committee for the Co-ordination and Standardisation of Speech Databases and Assessment Techniques, {O-COCOSDA} 2022, Hanoi, Vietnam, November 24-26, 2022}, pages 1--6. {IEEE}, 2022.

\bibitem{articleConcern}
Gold Okorie, Chioma Udeh, Ejuma Adaga, Obinna DaraOjimba, and Osato Oriekhoe.
\newblock Ethical considerations in data collection and analysis: A review: Investigating ethical practices and challenges in modern data collection and analysis.
\newblock {\em International Journal of Applied Research in Social Sciences}, 6:1--22, 01 2024.

\bibitem{Sasindran_2024}
Zitha Sasindran, Harsha Yelchuri, and T.~V. Prabhakar.
\newblock Semascore: A new evaluation metric for automatic speech recognition tasks.
\newblock In {\em Interspeech 2024}, interspeech\_2024, page 4558–4562. ISCA, September 2024.

\bibitem{roy2021semanticwerunifiedmetricevaluation}
Somnath Roy.
\newblock Semantic-wer: A unified metric for the evaluation of asr transcript for end usability, 2021.

\bibitem{chang2024interspeech2024challengespeech}
Xuankai Chang, Jiatong Shi, Jinchuan Tian, Yuning Wu, Yuxun Tang, Yihan Wu, Shinji Watanabe, Yossi Adi, Xie Chen, and Qin Jin.
\newblock The interspeech 2024 challenge on speech processing using discrete units, 2024.

\bibitem{10446418}
Linfeng Yu, Wangyou Zhang, Chenpeng Du, Leying Zhang, Zheng Liang, and Yanmin Qian.
\newblock Generation-based target speech extraction with speech discretization and vocoder.
\newblock In {\em ICASSP 2024 - 2024 IEEE International Conference on Acoustics, Speech and Signal Processing (ICASSP)}, pages 12612--12616, 2024.

\bibitem{poncelet2024efficientextractionnoiserobustdiscrete}
Jakob Poncelet, Yujun Wang, and Hugo~Van hamme.
\newblock Efficient extraction of noise-robust discrete units from self-supervised speech models, 2024.

\bibitem{hsu2021hubertselfsupervisedspeechrepresentation}
Wei-Ning Hsu, Benjamin Bolte, Yao-Hung~Hubert Tsai, Kushal Lakhotia, Ruslan Salakhutdinov, and Abdelrahman Mohamed.
\newblock Hubert: Self-supervised speech representation learning by masked prediction of hidden units, 2021.

\bibitem{Jin2010}
Xin Jin and Jiawei Han.
\newblock {\em K-Means Clustering}, pages 563--564.
\newblock Springer US, Boston, MA, 2010.

\bibitem{5453745}
Shi Na, Liu Xumin, and Guan Yong.
\newblock Research on k-means clustering algorithm: An improved k-means clustering algorithm.
\newblock In {\em 2010 Third International Symposium on Intelligent Information Technology and Security Informatics}, pages 63--67, 2010.

\bibitem{hasanabadi2023overviewtexttospeechsystemsmedia}
Mohammad~Reza Hasanabadi.
\newblock An overview of text-to-speech systems and media applications, 2023.

\bibitem{oord2018neuraldiscreterepresentationlearning}
Aaron van~den Oord, Oriol Vinyals, and Koray Kavukcuoglu.
\newblock Neural discrete representation learning, 2018.

\bibitem{chiu2022selfsupervisedlearningrandomprojectionquantizer}
Chung-Cheng Chiu, James Qin, Yu~Zhang, Jiahui Yu, and Yonghui Wu.
\newblock Self-supervised learning with random-projection quantizer for speech recognition, 2022.

\bibitem{randomQuantizationarticle}
Lahouari Cheded.
\newblock Random quantization: A new analysis with applications.
\newblock 07 2008.

\bibitem{whetten2024openimplementationstudybestrq}
Ryan Whetten, Titouan Parcollet, Marco Dinarelli, and Yannick Estève.
\newblock Open implementation and study of best-rq for speech processing, 2024.

\bibitem{astuti2022melweighted}
Yenni Astuti, Endang~Wulandari Utami, Latifatus~Sholikhah Aini, and Slamet~Widodo Prajitno.
\newblock A mel-weighted spectrogram feature extraction for improved speaker recognition system, 2022.
\newblock Accessed: 2025-01-01.

\bibitem{Zhang2019AudioRecognition}
Boyang Zhang, Jared Leitner, and Samuel Thornton.
\newblock Audio recognition using mel spectrograms and convolution neural networks.
\newblock In {\em Proceedings of the International Conference on Acoustics, Speech, and Signal Processing (ICASSP)}, 2019.
\newblock Accessed: 2025-01-01.

\bibitem{Boulal2024}
Hossam Boulal, Mohamed Hamidi, Mustapha Abarkan, and Jamal Barkani.
\newblock Amazigh cnn speech recognition system based on mel spectrogram feature extraction method.
\newblock {\em International Journal of Speech Technology}, 27(1):287--296, March 2024.

\bibitem{kong2020hifigangenerativeadversarialnetworks}
Jungil Kong, Jaehyeon Kim, and Jaekyoung Bae.
\newblock Hifi-gan: Generative adversarial networks for efficient and high fidelity speech synthesis, 2020.

\bibitem{mfcc2015}
Sayf~A. Majeed, Hafizah Husain, Salina~Abdul Samad, and Tariq~F. Idbeaa.
\newblock Mel frequency cepstral coefficients ({MFCC}) feature extraction enhancement in the application of speech recognition: A comparison study.
\newblock {\em Journal of Theoretical and Applied Information Technology}, 79(1), 2015.

\bibitem{articleMeteorScores}
Alon Lavie and Abhaya Agarwal.
\newblock Meteor: An automatic metric for mt evaluation with high levels of correlation with human judgments.
\newblock pages 228--231, 07 2007.

\bibitem{articleStft}
D.~Griffin and Jae Lim.
\newblock Signal estimation from modified short-time fourier transform.
\newblock {\em IEEE Transactions on Acoustics, Speech, and Signal Processing}, 32(2):236--243, 1984.

\bibitem{kent2002acoustic}
R.D. Kent and C.~Read.
\newblock {\em The Acoustic Analysis of Speech}.
\newblock Singular/Thomson Learning, 2002.

\bibitem{Griffin-Lim}
Yukoh Wakabayashi and Nobutaka Ono.
\newblock Griffin-lim phase reconstruction using short-time fourier transform with zero-padded frame analysis.
\newblock In {\em 2019 Asia-Pacific Signal and Information Processing Association Annual Summit and Conference (APSIPA ASC)}, pages 1863--1867, 2019.

\bibitem{oshea2015introductionconvolutionalneuralnetworks}
Keiron O'Shea and Ryan Nash.
\newblock An introduction to convolutional neural networks, 2015.

\bibitem{gulati2020conformerconvolutionaugmentedtransformerspeech}
Anmol Gulati, James Qin, Chung-Cheng Chiu, Niki Parmar, Yu~Zhang, Jiahui Yu, Wei Han, Shibo Wang, Zhengdong Zhang, Yonghui Wu, and Ruoming Pang.
\newblock Conformer: Convolution-augmented transformer for speech recognition, 2020.

\bibitem{joshi2020spanbertimprovingpretrainingrepresenting}
Mandar Joshi, Danqi Chen, Yinhan Liu, Daniel~S. Weld, Luke Zettlemoyer, and Omer Levy.
\newblock Spanbert: Improving pre-training by representing and predicting spans, 2020.

\bibitem{ilmt-s2s}
Akira Hayakawa, Saturnino Luz, Loredana Cerrato, and Nick Campbell.
\newblock The ilmt-s2s corpus ― a multimodal interlingual map task corpus.
\newblock In {Nicoletta Calzolari (Conference} Chair), Khalid Choukri, Thierry Declerck, Sara Goggi, Marko Grobelnik, Bente Maegaard, Joseph Mariani, Helene Mazo, Asuncion Moreno, Jan Odijk, and Stelios Piperidis, editors, {\em Proceedings of the Tenth International Conference on Language Resources and Evaluation (LREC 2016)}, page 605–612. European Language Resources Association (ELRA), May 2016.

\bibitem{Desplanques_2020}
Brecht Desplanques, Jenthe Thienpondt, and Kris Demuynck.
\newblock Ecapa-tdnn: Emphasized channel attention, propagation and aggregation in tdnn based speaker verification.
\newblock In {\em Interspeech 2020}, interspeech\_2020. ISCA, October 2020.

\bibitem{DBLP:journals/corr/VaswaniSPUJGKP17}
Ashish Vaswani, Noam Shazeer, Niki Parmar, Jakob Uszkoreit, Llion Jones, Aidan~N. Gomez, Lukasz Kaiser, and Illia Polosukhin.
\newblock Attention is all you need.
\newblock {\em CoRR}, abs/1706.03762, 2017.

\bibitem{DBLP:journals/corr/abs-2001-08210}
Yinhan Liu, Jiatao Gu, Naman Goyal, Xian Li, Sergey Edunov, Marjan Ghazvininejad, Mike Lewis, and Luke Zettlemoyer.
\newblock Multilingual denoising pre-training for neural machine translation.
\newblock {\em CoRR}, abs/2001.08210, 2020.

\bibitem{DBLP:journals/corr/abs-2006-09526}
Chau Tran, Yuqing Tang, Xian Li, and Jiatao Gu.
\newblock Cross-lingual retrieval for iterative self-supervised training.
\newblock {\em CoRR}, abs/2006.09526, 2020.

\bibitem{DBLP:journals/corr/abs-2103-15060}
Ye~Jia, Heiga Zen, Jonathan Shen, Yu~Zhang, and Yonghui Wu.
\newblock Png {BERT:} augmented {BERT} on phonemes and graphemes for neural {TTS}.
\newblock {\em CoRR}, abs/2103.15060, 2021.

\bibitem{shen2018naturalttssynthesisconditioning}
Jonathan Shen, Ruoming Pang, Ron~J. Weiss, Mike Schuster, Navdeep Jaitly, Zongheng Yang, Zhifeng Chen, Yu~Zhang, Yuxuan Wang, RJ~Skerry-Ryan, Rif~A. Saurous, Yannis Agiomyrgiannakis, and Yonghui Wu.
\newblock Natural tts synthesis by conditioning wavenet on mel spectrogram predictions, 2018.

\bibitem{oord2016wavenetgenerativemodelraw}
Aaron van~den Oord, Sander Dieleman, Heiga Zen, Karen Simonyan, Oriol Vinyals, Alex Graves, Nal Kalchbrenner, Andrew Senior, and Koray Kavukcuoglu.
\newblock Wavenet: A generative model for raw audio, 2016.

\bibitem{ren2022fastspeech2fasthighquality}
Yi~Ren, Chenxu Hu, Xu~Tan, Tao Qin, Sheng Zhao, Zhou Zhao, and Tie-Yan Liu.
\newblock Fastspeech 2: Fast and high-quality end-to-end text to speech, 2022.

\bibitem{kim2021conditionalvariationalautoencoderadversarial}
Jaehyeon Kim, Jungil Kong, and Juhee Son.
\newblock Conditional variational autoencoder with adversarial learning for end-to-end text-to-speech, 2021.

\bibitem{shi2022learningaudiovisualspeechrepresentation}
Bowen Shi, Wei-Ning Hsu, Kushal Lakhotia, and Abdelrahman Mohamed.
\newblock Learning audio-visual speech representation by masked multimodal cluster prediction, 2022.

\bibitem{chung2021w2vbertcombiningcontrastivelearning}
Yu-An Chung, Yu~Zhang, Wei Han, Chung-Cheng Chiu, James Qin, Ruoming Pang, and Yonghui Wu.
\newblock W2v-bert: Combining contrastive learning and masked language modeling for self-supervised speech pre-training, 2021.

\bibitem{song2023dspganganbaseduniversalvocoder}
Kun Song, Yongmao Zhang, Yi~Lei, Jian Cong, Hanzhao Li, Lei Xie, Gang He, and Jinfeng Bai.
\newblock Dspgan: a gan-based universal vocoder for high-fidelity tts by time-frequency domain supervision from dsp, 2023.

\bibitem{huang2022directedacyclictransformernonautoregressive}
Fei Huang, Hao Zhou, Yang Liu, Hang Li, and Minlie Huang.
\newblock Directed acyclic transformer for non-autoregressive machine translation, 2022.

\bibitem{bapna2022mslammassivelymultilingualjoint}
Ankur Bapna, Colin Cherry, Yu~Zhang, Ye~Jia, Melvin Johnson, Yong Cheng, Simran Khanuja, Jason Riesa, and Alexis Conneau.
\newblock mslam: Massively multilingual joint pre-training for speech and text, 2022.

\bibitem{GRAVES2005602}
Alex Graves and Jürgen Schmidhuber.
\newblock Framewise phoneme classification with bidirectional lstm and other neural network architectures.
\newblock {\em Neural Networks}, 18(5):602--610, 2005.
\newblock IJCNN 2005.

\bibitem{Park_2019}
Daniel~S. Park, William Chan, Yu~Zhang, Chung-Cheng Chiu, Barret Zoph, Ekin~D. Cubuk, and Quoc~V. Le.
\newblock Specaugment: A simple data augmentation method for automatic speech recognition.
\newblock In {\em Interspeech 2019}, interspeech\_2019. ISCA, September 2019.

\bibitem{58337}
P.J. Werbos.
\newblock Backpropagation through time: what it does and how to do it.
\newblock {\em Proceedings of the IEEE}, 78(10):1550--1560, 1990.

\bibitem{caruana1997multitask}
Rich Caruana.
\newblock Multitask learning.
\newblock {\em Machine Learning}, 28(1):41--75, 1997.

\bibitem{sarkar2023xkdcrossmodalknowledgedistillation}
Pritam Sarkar and Ali Etemad.
\newblock Xkd: Cross-modal knowledge distillation with domain alignment for video representation learning, 2023.

\bibitem{lester2021powerscaleparameterefficientprompt}
Brian Lester, Rami Al-Rfou, and Noah Constant.
\newblock The power of scale for parameter-efficient prompt tuning, 2021.

\bibitem{hsu2023pseudolabeltrainingmodelinertia}
Benjamin Hsu, Anna Currey, Xing Niu, Maria Nădejde, and Georgiana Dinu.
\newblock Pseudo-label training and model inertia in neural machine translation, 2023.

\bibitem{germain2015mademaskedautoencoderdistribution}
Mathieu Germain, Karol Gregor, Iain Murray, and Hugo Larochelle.
\newblock Made: Masked autoencoder for distribution estimation, 2015.

\end{thebibliography}

\end{document}